%% file: main.tex
\pdfoutput=1

\documentclass[11pt]{article}

\usepackage[]{acl}

\usepackage{times}
\usepackage{latexsym}

\usepackage[T1]{fontenc}

\usepackage[utf8]{inputenc}
\usepackage[compact]{titlesec}

\usepackage{microtype}

\usepackage{inconsolata}
\usepackage{enumitem}
\usepackage{subcaption}

\usepackage{booktabs}
\usepackage{amssymb}
\usepackage{graphicx}

\usepackage{multirow}

\usepackage{longtable}
\usepackage{xspace}
\usepackage{xurl}
\usepackage{svg}

\newcommand{\cv}{\texttt{CV}\xspace}
\newcommand{\voxpopuli}{\texttt{VoxPopuli}\xspace}
\newcommand{\fleurs}{\texttt{Fleurs}\xspace}
\newcommand{\whisper}{\textsc{Whisper}\xspace}
\newcommand{\seamless}{\textsc{Seamless}\xspace}

\newcommand{\repo}{\url{https://github.com/g8a9/multilingual-asr-gender-gap}\xspace}

\title{Twists, Humps, and Pebbles: Multilingual Speech Recognition Models Exhibit Gender Performance Gaps}

\newcommand{\bocconi}{$^{\clubsuit}$}
\newcommand{\fbk}{$^{\spadesuit}$}
\newcommand{\utn}{$^{\diamondsuit}$}
\newcommand{\lis}{$^{\heartsuit}$}

\author{Giuseppe Attanasio\lis, Beatrice Savoldi\fbk, Dennis Fucci\utn\fbk, Dirk Hovy\bocconi \\ \\
 \lis~Instituto de Telecomunicações, Lisbon, Portugal \\
 \utn~University of Trento, Trento, Italy\\
 \fbk~Fondazione Bruno Kessler, Trento, Italy\\
 \bocconi~Bocconi University, Milan, Italy \\
    \texttt{\href{mailto:giuseppe.attanasio3@unibocconi.it}{giuseppe.attanasio@lx.it.pt}}
} 

\begin{document}
\maketitle
\begin{abstract}

Current automatic speech recognition (ASR) models are designed to be used across many languages and tasks without substantial changes.
However, this broad language coverage hides performance gaps \textit{within} languages, for example, across genders.
Our study systematically evaluates the performance of two widely used multilingual ASR models on three datasets, encompassing 19 languages from eight language families and two speaking conditions. Our findings reveal clear gender disparities, with the advantaged group varying across languages and models.
Surprisingly, those gaps are not explained by acoustic or lexical properties.
However, probing internal model states reveals a correlation with gendered performance gap.
That is,
the easier it is to distinguish speaker gender in a language using probes, the more the gap reduces, favoring female speakers.
Our results show that gender disparities persist even in state-of-the-art models.
Our findings have implications for the improvement of multilingual ASR systems, underscoring the importance of accessibility to training data and nuanced evaluation to predict and mitigate gender gaps.  
We release all code and artifacts at \repo.

\end{abstract}

\section{Introduction}

\begin{figure}[!t]
    \centering
    \includegraphics[width=\linewidth]{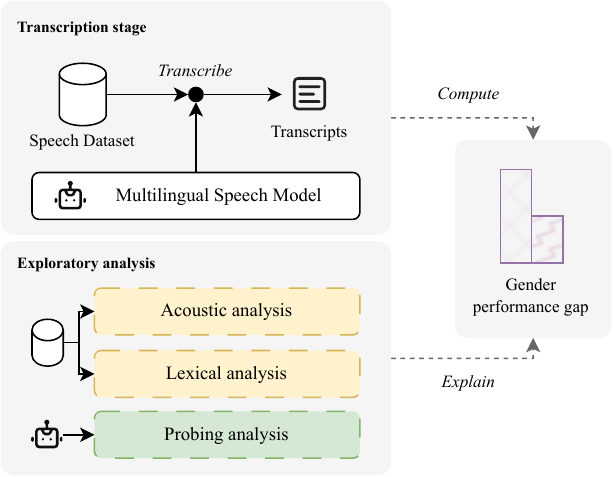}
    \caption{\textbf{Study overview.}
    We transcribe three speech datasets across 19 languages and compute gender performance gaps. Next, 
    we
    investigate the models for possible causes of these gaps.}
    \label{fig:summary}
\end{figure}

A new class of multi-task, multilingual neural networks \citep{radford_robust_2022,communication2023seamless,chu2023qwen} has recently pushed the boundaries of several speech-related tasks, including automatic speech recognition (ASR). As these models offer support to an increasing number of languages at no cost, they have found widespread adoption and have been integrated into applications for the general public, such as real-time voice transcription.\footnote{See, for instance, \url{http://tcrn.ch/4el8Brr}} However, while the user base expands, one question has yet to be answered: \textit{will usability be the same for everyone?}
In this paper, we seek an answer to this question by studying how systems understand the voices of different genders and across multiple languages. These two axes of analysis are motivated by complementary interests.

On the one hand, voice and language production are among the strongest identity traits. They vary 
across individuals, sociodemographic groups \citep{labov1964social, wolfram2004urban,alim2004you}, and, crucially, across genders. 
Gendered differences in the voice---rooted in physiological factors linked to biological sex (e.g., vocal tract length), as well as in sociocultural ones (e.g., prescribed registers) \citep{zimman2017gender,zimman2020sociophonetics}---have been extensively studied in sociolinguistics \cite[e.g.,][]{coleman-1976-voice,busby1995formant,hillenbrand-2009-pitch}. 
Also, empirical studies have shown that gender affects the performance of traditional ASR systems in English \citep{garnerin-etal-2019-gender}. 
On the other hand, voice and speech can change drastically 
across languages, cultures, \textit{and} in function of gender. 
For example, males' and females' pitch---a measure of the highness of the voice---is much closer in Japanese- than in English-speaking people \citep{loveday1981pitch, yuasa2008culture}.  

Despite a large literature on gender, voice, and ASR, to our knowledge, 
no studies have tested whether multi-task, multilingual ASR models serve genders to a comparable level across languages. Arguably, overlooking potential disparities can result in unequal service quality for already socially disadvantaged individuals \citep{mengesha2021, tatman-2017-gender}. 

To fill this gap, we systematically study gendered performance gaps in massively multilingual ASR models. We set out to answer two research questions. 
(\textbf{Q1})\textit{ Do multi-task, multilingual ASR models perform equally well across speakers identifying as women, men, or neither of the two?}
%
If so, (\textbf{Q2}),\textit{ can we relate gender gaps to acoustic and lexical variation in data or model internal states?}


To answer Q1, we evaluate two state-of-the-art multilingual open-weight ASR models on three datasets, 
covering 19 languages and two speech conditions (i.e. read and spontaneous).
The results of our extensive evaluation show that \textbf{models systematically exhibit gender performance gaps}.
However, whether models favor male or female speakers depends on the dataset and language.
Results on one dataset also highlight subpar performance for speakers who do not identify with either gender compared to males.

While studying acoustic and lexical phenomena in test data (Q2), we found no significant correlations of specific features with performance gaps between male and female speakers. This objective finding underscores the complexity of the issue and the need for further research.  
Instead, interpretability analyses suggest that when presented with speech from men and women, models build different internal representations that can serve as a proxy for gender disparities.


\paragraph{Contributions.}
We conduct the first extensive evaluation of two widely used multilingual ASR models for gender performance differences. We document significant gender gaps by inspecting model internal states. 
We release code, data, and all artifacts we produce for future research.

\paragraph{Bias Statement.}
When using gender as a variable, we rely on  
speakers' declared identity (see \S\ref{sec:ethics}). We evaluate whether speech from individuals identifying as female, male, or neither of the two is processed equally well by multilingual ASR models. 
We measure disparities of minority groups versus the socially advantaged group, i.e., men. Performance parity is the ideal outcome.
We define a system to be biased if it risks further contributing allocative (i.e., quality of service, technology less accessible) and representational harms impacting the minority groups, e.g., feeding into stereotypes about the inadequacy of women and speech technology or ``shrill,'' ``incorrect'' voice \citep{Tallon}.\footnote{As voice technology expert Tom Schalk once put it, ``many issues with women's voices could be fixed if female drivers were willing to sit through lengthy training... Women could be taught to speak louder and direct their voices towards the microphone.'' \url{https://bit.ly/time-shalck} }

\section{Background}
\label{sec:background}

\paragraph{Multi-task Multilingual Speech Models.}

Contemporary speech recognition systems process audio and text separately and use common cross-modal interactions---e.g., OpenAI's Whisper \citep{radford_robust_2022} is loosely 
inspired by the
standard Transformer \cite{vaswani2017attention}.
Crucially, multitasking and multilinguality come at virtually no cost for the user. With Whisper, for example, special ``task'' and ``language'' tokens can be prepended to the decoder input to change its functions.   
This strategy allows multi-tasking and multilinguality without architectural changes or fine-tuning (see Figure 1, top-right, in \citet{radford_robust_2022}).
Similarly, Meta's SeamlessM4T \citep{communication2023seamless} uses a speech encoder for audio and an encoder-decoder transformer for text \citep[NLLB]{costa2022no}.



\paragraph{Gender and Speech (Technologies).}
Gendered aspects of the voice are one of the most salient individual traits \citep{kreiman2011foundations, azul2015varied, zimman2021gender} and have long been studied in linguistics \citep{zimman2020sociophonetics}. 
The anatomy and makeup of the vocal tract do play a role, as it determines pitch range and formant variations, often regarded as the most distinctive 
vocal features of cisgender men and women.\footnote{\textit{Cisgender} describes individuals whose gender identity matches their birth-assigned sex \citep{fuchs2010differences}.}
However, gender variation in the voice 
has been shown to 
arise from several aspects besides physical ones \citep{oates2015transgender, zimman2018transgender, Becker_Khan_Zimman_2022}.
As studies including transgender individuals show \cite{zimman2017gender}, sociocultural factors influence vocal use, including which parts of the pitch range are used \citep{loveday1981pitch, yuasa2008culture}, and articulatory practices for sibilant consonants and vowels that are perceived as gender characterizing \citep{pharao2014, podesva2016s, li2017development}.

For speech technologies, sociodemographic variation has posed challenges to ASR systems \citep{sawalha2013effects, liu-etal-2022-female,rajan2022aequevox, fucci2023no}. 
Various works have found ASR models for English and French to recognize better male speech and voices \citep{tatman-2017-gender, garnerin-etal-2019-gender, garnerin-etal-2021-gender}. This effect is often a result of the under-representation of women in the training data \citep{meyer2020artie, gaido-etal-2020-breeding, garnerin-etal-2020-gender}. 
However, various works have found the reverse to be true. Several studies found comparatively better performance for women in more spontaneous, conversational data in Dutch \citep{feng-etal-2022-ceres} and English \citep{addadecker05_interspeech, koenecke2020}. They attribute these findings to sociolinguistic factors, like the higher incidence of disfluencies and informal speech in men. 


Thus, the current literature provides a fragmented picture of gender disparities in speech technologies. However, most studies focus primarily on a single language, usually English, albeit across multiple datasets and models. 
We broaden this research to include large-scale \textit{multi}lingual speech models. 
In addition, we evaluate performance in a third gender category, which includes individuals who do not identify as male or female.

\section{Methodology}
\label{ssec:q1_exps}

This section describes the experimental design to answer \textbf{Q1}, i.e., whether there are gendered performance differences in multi-task multilingual ASR models. 
Our results (\S\ref{ssec:q1_results}) confirm it.


\paragraph{Models.} We experiment with OpenAI's Whisper \citep[\whisper]{radford_robust_2022} and Meta's SeamlessM4T \citep[\seamless]{communication2023seamless}, two widely used, state-of-the-art multilingual ASR models (details in Appendix \ref{app:models}).



\paragraph{Datasets.} Among other multilingual datasets available for ASR \citep{gales2014speech, CMU-2019, MLS-2020,iranzo-sanchez-etal-2022-mllp, valk2021voxlingua107}, we use Mozilla Common Voice \citep[\cv,][]{ardila2020common},\footnote{We use CV 16.0 from \url{https://huggingface.co/datasets/mozilla-foundation/common_voice_16_0}} \texttt{Fleurs} \citep{conneau2023fleurs} and \voxpopuli \citep{wang-etal-2021-voxpopuli}. Mainly, dataset selection is bound to the availability of reliable speakers' gender information. We use the gender labels that come with each dataset.\footnote{Speakers indicated gender identity with ``Male'' or ``Female'' labels in \fleurs and \cv. The latter also includes an ``Other'' label. Gender is retrieved from \url{https://multimedia.europarl.europa.eu/en} in \voxpopuli.}
Moreover, these datasets cover two distinct recording and speech conditions: \textit{i) read}, where speech is typically well-articulated and based on the reading of pre-defined texts (\cv and \fleurs), and \textit{ii)} \textit{spontaneous} conditions elicited from public speeches (\voxpopuli), which allow for more speaker-dependent articulations and variation in word usage \citep{gabler2023reconsidering}, and thus represent a testbed closer to real-world use cases of ASR technologies.  
We base our analysis on the concatenation of validation and test splits of each dataset to avoid unreliable results due to training data contamination. See Appendix~\ref{app:sampling_effect} for a discussion on data contamination and how transcription performance varies across splits in our setup.

%

\paragraph{Languages.} We include 19 languages from \cv and \fleurs, and a subset of those---11 in total, due to data availability---from \voxpopuli. They represent eight diverse language families and data availability conditions (i.e., high-low resource). The dataset choice depended on whether sufficient utterances stratified across genders were available for meaningful comparisons. See Tables~\ref{tab:stats_cv}-\ref{tab:stats_fleurs}, Appendix \ref{app:dataset_statistics} for all the statistics.

\subsection{Evaluation}
\label{sec:eval}

\paragraph{Quality Metrics.}
We use standard Word- and Character-Error Rate (WER and CER, respectively) to evaluate transcription quality. Following \citet{radford_robust_2022}, we 
\textit{i}) report WER for all languages but Yoruba and Japanese---where variability in orthography and ambiguous word units may affect evaluation \cite{rowlands1954types,matsuoka1997japanese}---for which we use CER, 
\textit{ii}) transliterate all Russian and Serbian references and hypotheses into Cyrillic, and
\textit{iii}) apply the official normalization routines to texts when evaluating Whisper.\footnote{Found at \url{https://github.com/openai/whisper/tree/main/whisper/normalizers}}
Moreover, we trim audio to the initial 30 seconds\footnote{Most recordings in our datasets are shorter than 30 seconds. See statistics in Table~\ref{tab:stats_cv}-\ref{tab:stats_fleurs} in Appendix~\ref{app:dataset_statistics}.}
and filter out dataset noise by removing records with an empty reference and silence in the snippet.\footnote{When processing \texttt{Fleurs-es}, we noticed some records containing only background noise. We used state-of-the-art voice-activity detection models to detect and filter them out. See Appendix~\ref{app:vad} for details.}

When evaluated in in terms of 
of overall quality, both
\whisper and \seamless showcase competitive results, respectively, on \texttt{Fleurs} (12.68 and 19.87 -- avg. 19 langs); \texttt{CV} (17.51 and 16.42 -- avg. 19 langs); \texttt{VP} (13.59 and 12.74 -- avg. 11 langs). For disaggregated error rate (ER) scores, see Figure \ref{fig:wer} in Appendix \ref{app:models}.

\paragraph{Gap Metrics.}

Measuring the ASR gender gap is equivalent to seeking what is commonly known in the field of machine learning as \textit{group fairness} \citep{chouldechova2020snapshot}, or, more specifically, \textit{demographic parity} \citep{dwork2012fairness}. Conceptually, parity is achieved when a given statistical measure is equal across different groups. 

We operationalize demographic parity using the notation from \citet{czarnowska-etal-2021-quantifying}. We introduce a Pairwise Comparison Metric, loosely inspired by the Disparity Score in \citet{gaut-etal-2020-towards}, defined as follows:
\begin{equation} \label{eq:gap_metric}
    E(r_A,r_B) = 100 \cdot (\phi(r_A) - \phi(r_B)) / \phi(r_B)
\end{equation}

\noindent where $r_A$ and $r_B$ are the set of audio snippets belonging to given groups (e.g., either ``male'', ``female'', or ``other'' in \cv), and $\phi(\cdot)$ is one of the quality metrics defined above (i.e., WER or CER). The ideal score is 0---i.e., the model performs perfectly equally on the two groups.\footnote{ASR models achieve different baseline error rates across languages. Equation~\ref{eq:gap_metric} measures relative, rather than absolute, gaps to account for this variability. See Table~\ref{tab:wer_diff_splits} in Appendix~\ref{app:sampling_effect} for details on absolute gaps and \S\ref{sec:discussion} for a discussion on possible metric interpretations.}

\paragraph{Significance test.}





We use a bootstrapped approach \citep{koehn-2004-statistical,sogaard-etal-2014-whats} to estimate $\phi(\cdot)$.
We sample $40\%$ of the smallest group and the same number of records from the largest group for $n=1000$ iterations. We sample stratifying on speakers to avoid skewing the speaker distribution. Stratified, gender-balanced sampling ensures a reliable comparison. 
We compute $E(\cdot)$ on the arithmetic means of $\phi(\cdot)$ across the $n$ runs and use a two-sided Student's t-test to compute the statistical difference between the means.
See Appendix~\ref{app:sampling_effect} for more details on comparing sampling performance vs. complete sets.

Note that sampling requires attributing each record to an individual speaker---information not available in \fleurs.
For consistency, we attribute a speaker ID to each record automatically. Specifically, we \textit{i}) use state-of-the-art speaker verification embedding models to encode each recording, \textit{ii}) cluster them using HDBSCAN \citep{campello2013density}, and \textit{iii}) assign a speaker ID to each cluster. See Appendix~\ref{app:fleurs_hdbscan} for details on the pipeline.





\section{Gender Performance Gaps}
\label{ssec:q1_results}
In the following, we report the results of evaluating \whisper and \seamless, focusing on gender performance disparities. We first investigate whether models are equally able to recognize female and male speakers (\S\ref{subsec:female_male_gap}). Then, limited to \cv, we repeat the analysis for speakers identifying with neither of the two against males (\S\ref{subsec:other_male_gap}).   





\begin{figure}[!t]
    \centering
    \begin{subfigure}{\columnwidth}
        \centering
        \includegraphics[width=\linewidth]{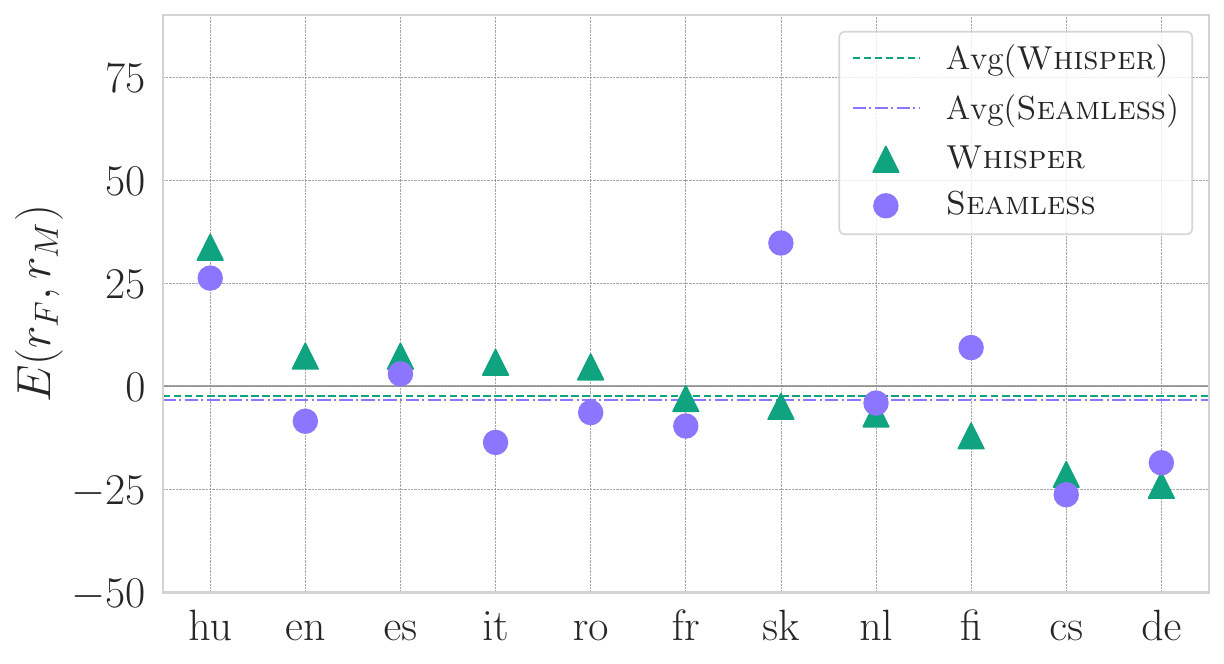}
        \caption{\voxpopuli}
        \label{fig:VP}
    \end{subfigure}
    
    \begin{subfigure}{\columnwidth}
        \centering
        \includegraphics[width=\linewidth]{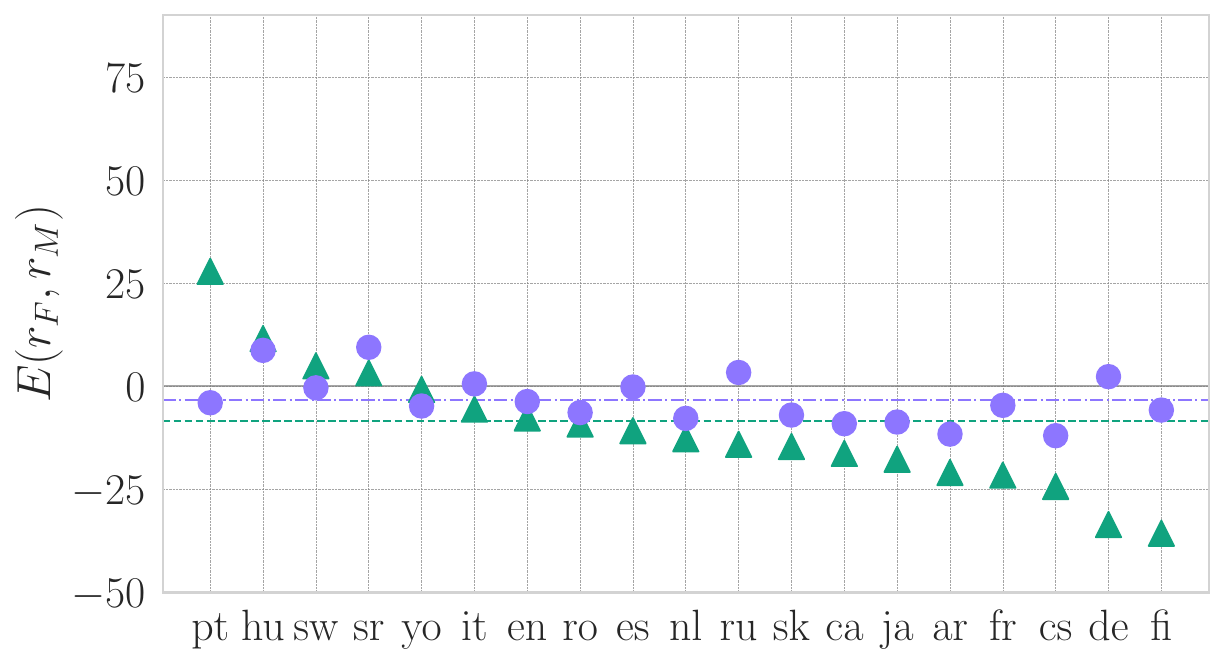}
        \caption{\fleurs}
        \label{fig:fleurs}
    \end{subfigure}
    
    \begin{subfigure}{1\columnwidth}
        \centering
        \includegraphics[width=\linewidth]{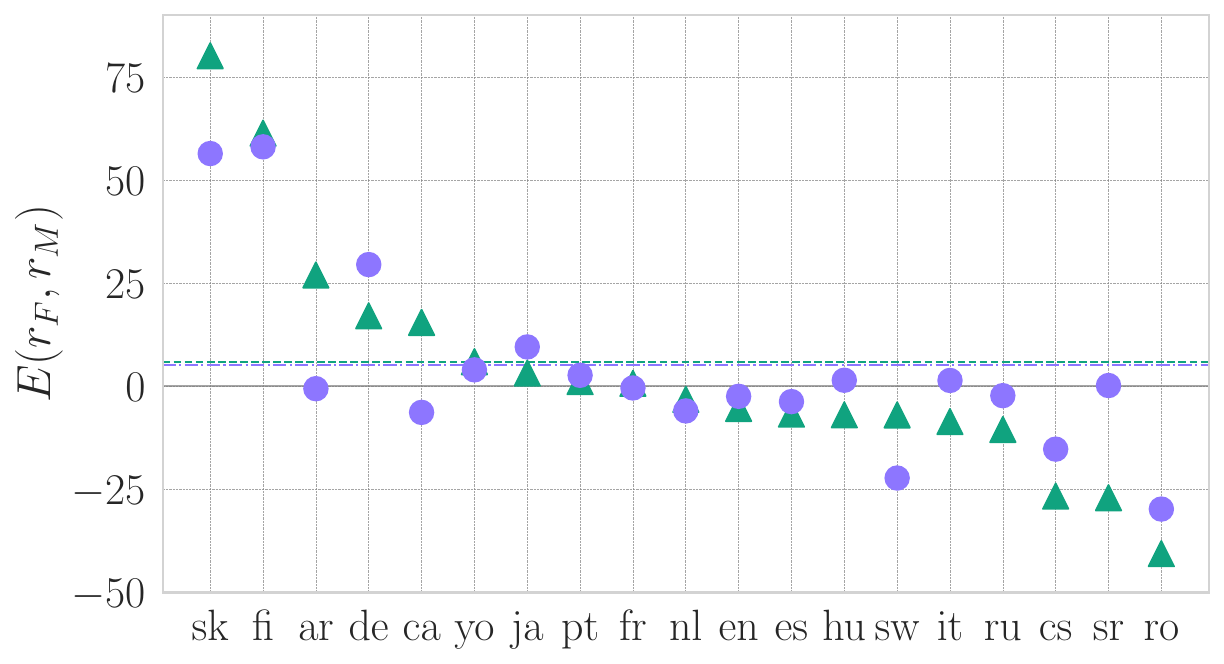}
        \caption{\cv}
        \label{fig:CV}
    \end{subfigure}
    \caption{\textbf{Model error rate gap (Eq. \ref{eq:gap_metric}).} Positive values indicate better performance on men, negative ones on women. $p<0.05$.}
    \label{fig:FM_rel_diff}
\end{figure}

\subsection{``Female'' -- ``Male'' Gap}
\label{subsec:female_male_gap}


Figure \ref{fig:FM_rel_diff} reports $E(r_F,r_M)$, i.e., the gender gap results for \whisper and \seamless on each dataset. 
Concerning our \textbf{Q1}, a broad overview reveals that these models do not perform equally across female and male speakers, often showing a preference for one gender over the other. This gender disparity in multilingual models is a key point of our investigation.
Our analysis does not consistently reveal models disadvantaging the feminine group---see values below 0. This finding stands in stark 
contrast with mounting evidence of a strong masculine bias affecting a wide range of Natural Language Processing (NLP) tasks \citep[e.g.,][]{sun-etal-2019-mitigating,stanczak-etal-2022-neurons}. It is especially noteworthy given the well-known under-representation of women in current resources used for model training \citep{garnerin-etal-2019-gender, zanon-boito-etal-2022-speech, sun2021men}. We observe that both speech models---albeit to varying degrees---%
better recognize female speech on average in two out of the three considered datasets (i.e., \texttt{Fleurs} and \texttt{VP}).

Upon closer examination, we also observe substantially different behaviors across datasets. First, despite the higher degree of spoken language variation to be expected in \textit{spontaneous} recordings from the \texttt{VP} dataset (Figure \ref{fig:VP}), results shows a comparatively reduced gender gap, with overlapping values between \whisper and \seamless for most languages (i.e., hu, es, fr, nl, cs, de). Conversely, it is on \textit{read} resource that we attest to a higher degree of variability across models as well as languages, with occasional wide 
disparities toward either females or males in \cv (\ref{fig:CV}). Notably, \seamless remains closer to performance equality for most instances on \fleurs (\ref{fig:fleurs}), i.e., values between 11 and -10.

As such, in line with previous research (see \S\ref{sec:background}), analysis of (binary) gender disparities for the ASR reveals a complex picture, which we further analyze and unpack in the following sections.

\subsection{``Other'' -- ``Male'' Gap}
\label{subsec:other_male_gap}


Restricting our analysis to validation and test sets leaves us with few records for speakers who do not identify as either male or female. As such, we report results for five high-resource languages, Catalan, German, English, Spanish, and French.\footnote{Size varies from 8 
(\texttt{ca}, min) to 86 (\texttt{es}, max) records.}

Table~\ref{tab:other_male_gaps} reports $E(r_O,r_M)$ for both models.
Results suggest that models penalize non-male speech and that gaps are larger than disparities observed for the female/male groups.
Given the limited sample size, we deem this analysis only exploratory. However, it is a valuable first overview of the model's behaviors for gender-non-conforming voices.
We underscore the need for more representation of diverse voices and sociodemographic groups in current resources, and for which monitoring fairness in existing 
speech models
remains out of reach. See \S\ref{sec:discussion} for an expanded discussion.


\section{Acoustic and Lexical Analysis}
\label{sec:acoustic}




Gendered performance gaps vary across languages and datasets.
Here, we dive into a focused analysis on acoustic aspects or lexical phenomena present in the test data (\textbf{Q2}). Due to data availability, we conduct this analysis on the ``Female'' - ``Male'' setup.

%
%
%

\subsection{Acoustic Analysis}
\label{ssec:acoustic_analysis}

\begin{table}[!t]
\footnotesize
\centering
\begin{tabular}{@{}lccccc@{}}
\toprule
\textbf{} & \textbf{ca} & \textbf{de} & \textbf{en} & \textbf{es} & \textbf{fr} \\ \midrule
\whisper & 4.68 &
37.20 &
28.50 &
-2.00 &
5.21 \\
\seamless & 52.76 &
17.19 & 
38.70 &
11.37 &
-0.50   \\ \bottomrule
\end{tabular}
\caption{\textbf{Model error rate gap (Eq. \ref{eq:gap_metric}).} Positive values indicate better performance on men, negative ones on ``Other.'' $p < 0.05$.}
\label{tab:other_male_gaps}
\end{table}

Motivated by sociophonetic evidence for gendered differences in speech (\S\ref{sec:background}), we examine voice and language production across speakers of different genders. Specifically, we measure three acoustic features in our evaluation records, and explore whether potential acoustic differences across gender groups relate to gender performance gaps.

%
We include:
\textit{i}) \textit{pitch}, measured as the mean of the fundamental frequency values \cite{hirst-2021-f0}, known to vary between biological sexes \cite{coleman-1976-voice, hillenbrand-2009-pitch} and languages \citep{loveday1981pitch, yuasa2008culture};
\textit{ii}) \textit{intensity}, measured as the mean of fundamental frequencies \cite{Pausewang-1997-intensity}, subject to recording conditions \cite{youri-2015-recording};
\textit{iii}) \textit{speaking rate}, measured as the number of tokens per minute \cite{Knzel2013SomeGP}, subject to language and gender variation \citep{vanborsel-2008-rate,coupe-etal-2019-fast} as well as read vs. spontaneous speech \cite{nakamura-etal-2008-spontaneous}.\footnote{We compute average pitch and intensity using Praat \cite{praat}. Concerning speaking rate, we found no reliable multilingual tools for syllable segmentation. Therefore, we approximated it by the number of tokens as computed by \whisper's \href{https://huggingface.co/openai/whisper-large-v3/blob/main/tokenizer_config.json}{tokenizer}.}


\paragraph{Findings.}

We started by exploring whether pitch, intensity, and speaking rate differ between genders in each dataset-language pair. 
Independent-sample T-tests revealed significant differences for most setups ($p<0.05$), consistent with previous literature. For example, pitch consistently showed statistically significant differences (see 
Appendix~\ref{app:ssec:acoustic_analysis}).

Motivated by these observations, we computed the mean acoustic values of each gender group and correlated it with performance gaps, i.e., $E(r_F,r_M)$, aggregating by dataset and model.
No clear trends emerged overall. Sporadically, we found a strong and significant linear correlation, e.g., for \seamless and \voxpopuli where Pearson's \textit{rho} for pitch is $-0.83$ (see Appendix~\ref{app:ssec:acoustic_analysis} for complete details).
%
Fitting an ordinary least squares (OLS) regressor to predict $E(r_F,r_M)$ led to similar conclusions, as low R$^2$ scores suggest (max: $0.42$, average$_\sigma$: $0.24_{\pm0.11}$). For a more fine-grained analysis, we repeated the analysis fitting an OLS model to predict sentence-level error rates (i.e., $r_F$, $r_M$) but found low R$^2$ scores (max: $0.20$, average$_\sigma$: $0.03_{\pm0.04}$).



In summary, despite pitch, intensity, and speaking rate vary significantly across gender groups within the same dataset and language, such variation seldom correlates with performance differences. These findings suggest that performance gaps are a complex phenomenon, and our analysis should expand elsewhere to explain it.

\subsection{Lexical Analysis}

In \S\ref{ssec:acoustic_analysis}, we 
explored acoustic aspects of speech. 
However, \textit{what} is uttered can be as crucial as \textit{how} it is uttered.
Indeed, prior work has found that overall speech perception can be impacted by aspects related to lexical and syntactic complexities \citep{vanKnijff2018SpeechUI,Carroll2012TheEO}. 
Besides, certain lexical phenomena such as named entities represent a well-known challenge for speech models, especially in multilingual contexts \citep{gaido2021moby}.
Thus, we focus on speech content and study whether lexical phenomena---as measured in the reference transcripts---explain ASR disparities.


For each record, we counted \textit{i}) the occurrences of part-of-speech tags and \textit{ii}) named entities, and computed \textit{iii}) lexical density \citep{Halliday1989SpokenAW}.
Similarly to acoustic features, we contrast distribution between gender groups (details in Appendix~\ref{app:lexical_analysis}).

\paragraph{Findings.} Comparing the distribution of lexical features between female and male speakers, we found mixed results. 
Several dataset-language setups have significantly different distributions ($p<0.05$), but not all. 
Different distributions are also present in read datasets, suggesting that lexical variability is not controlled when collecting data.
We proceeded similarly to the acoustic analysis to verify whether such differences explain error rates.
Fitting an OLS model to predict error rates from lexical features only yields low R$^2$ scores (max: 0.35, average$_\sigma$: 0.09$_{\pm0.08}$, across all datasets and languages). Moreover, we found no significant (linear) correlation between the difference of group means and $E(r_F,r_M)$. 

This finding echoes those from acoustic analysis: Lexical phenomena do not explain gender performance gaps in our data. We must research other aspects beyond the lexical content of utterances.

\section{Probing Gender in ASR Models}





The field of natural language processing (NLP) has now established that transformer language models encode syntactic \citep{hewitt-manning-2019-structural}, semantic \citep{tenney-etal-2019-bert}, and factual \citep{petroni-etal-2019-language,meng2022locating} information in hidden representations. As such, recent work has focused on \textit{extracting} this information through probes, i.e., supervised classifiers trained on the model's embeddings \citep{alain2016understanding,belinkov-glass-2019-analysis}.
Compellingly, some have related extractability of sensitive attributes, e.g., gender, to bias in downstream application \citep{orgad-etal-2022-gender}. Probing is primarily motivated by the risk that models entangle protected attributes and predictions in sensitive use cases \citep{zhao-etal-2018-learning,ravfogel-etal-2020-null}. 

Motivated by this line of research, we measure gender extractability in ASR models and ask (\textbf{Q2}) whether and to what extent it explains gender-based performance gaps. To our knowledge, ours is the first study of this kind.\footnote{Interpretability for speech models is a relatively new research avenue. \citet{mohebbi-etal-2023-homophone} use context-mixing techniques to explain homophone disambiguation; \citet{pastor2023explaining} mask word-units to explain intent detection models. Following our work, \citet{krishnan2024encodinggendertransformerbasedasr} use amnesic probing to linearly erase gender information in transformer-based ASR models, examining its effects on downstream performance.} 

\paragraph{Experimental design.} We focus on one dataset-model configuration, namely \whisper and \cv. We attach our probes to the model encoder's last layer embeddings and train one distinct probe for every position. 
We use Logistic Regression and Minimum Description Length (MDL) \citep{voita-titov-2020-information} to probe gender in the female-male binary setup (see Appendix~\ref{app:probing_details} for details).

\begin{figure}[!t]
    \centering
    \includegraphics[width=\linewidth]{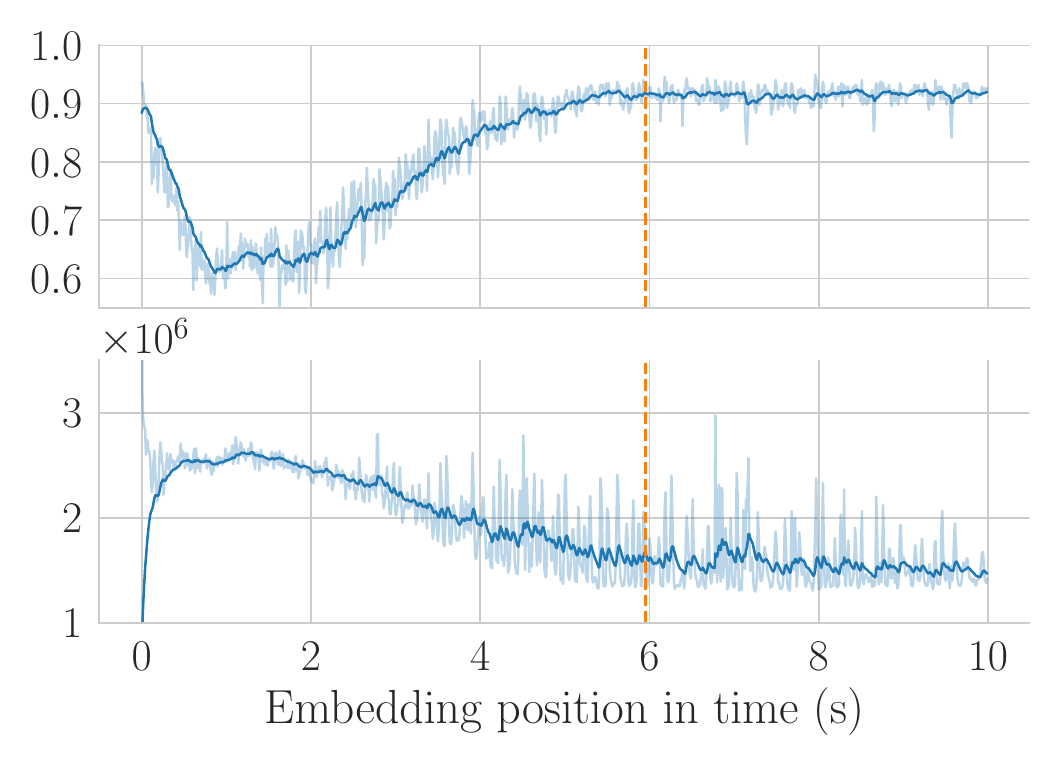}
    \caption{\textbf{F1 Macro (top) and measured code length (bottom)} for Logistic Regression and MDL probes, respectively. English, F-M setup, first 10 seconds of context. Actual score (blue pale line), exponential moving average (solid line, n=3), and average length of test snippets (dashed line).}
    \label{fig:compression}
\end{figure}

\paragraph{Findings.}

Figure~\ref{fig:compression} shows gender extractability on the English \cv split for the F-M setup. Trends are similar across languages (full results in Figure~\ref{sfig:logreg_fm_normal}).
First, F1 scores suggest that \textbf{gender extractability is relatively easy}. Whisper produces representations of female and male speech that can be easily separated.
Second, extractability is not constant over time. It starts high in the first milliseconds, drops during the actual signal (i.e., the speaker is talking), and finally plateaus around the initial value. This finding recalls the ``attention sink'' theory, implicating that transformers prioritize initial positions for specific tasks \cite{xiao2023efficient}. 
Third, the trends observed in MDL probes---where lower codelength indicates an easier task and higher extractability---suggest that simple logistic regression is a robust method, countering the issues identified in NLP
\citep{voita-titov-2020-information}.\footnote{To confirm this hypothesis, we conduct additional experiments by training logistic probes with randomly shifted labels. We report random guess performance on all languages (see Figure~\ref{sfig:logreg_fm_random}) in Appendix~\ref{app:probing_details}.}



\paragraph{Correlation with Error Rate.}

Do gender probing scores tell us something about ASR quality and gender bias (\textbf{Q2})?
To answer this question, we compare F1 scores from logistic regression scores and ASR error rates.
We measure F1$_S$, i.e., the average F1 score achieved by probes at the positions corresponding to actual speech,\footnote{We approximate this interval by considering all positions before the average test set length (orange line in Figure~\ref{fig:compression}).} $\phi(r_M)$ and $\phi(r_F)$ (error rates on male and female segments), and the disparity score $E(r_F,r_M)$.

\begin{figure}[!t]
    \centering
    \includegraphics[width=\linewidth]{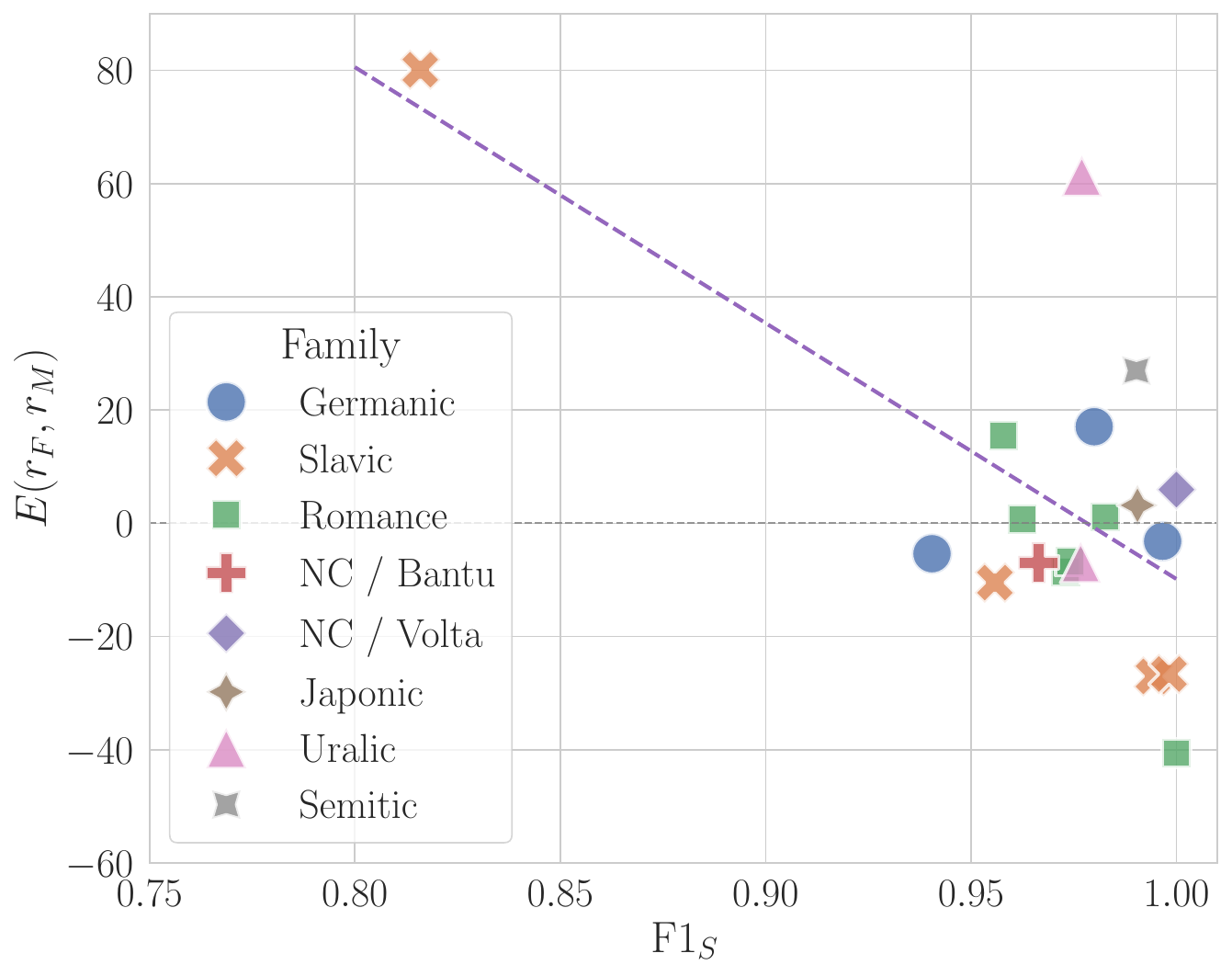}
    \caption{\textbf{Logistic probe F1$_S$ vs. $E(r_F,r_M)$}. \cv. The purple dashed line is a linear interpolation.}
    \label{fig:pearson}
\end{figure}

F1$_S$ has a weak linear correlation with $\phi(r_M)$ and $\phi(r_F)$ (Pearson's $\rho$ is 0.13 and -0.003, respectively), hinting that F1$_S$ cannot explain per-group quality. 
However, $\rho$ between F1$_S$ and $E(r_F,r_M)$ is -0.65 ($n=19$, see Figure~\ref{fig:pearson}). In other words, linear correlation suggests that \textbf{the better probes can extract gender from \whisper's hidden states, the lower the F-M error rate gap is}. Note that, as per its definition, $E(r_F,r_F)$ can be negative (see points below the zero line in Figure~\ref{fig:pearson}). Negative values indicate performance favoring the minority group (here, women).
This finding suggests that \whisper \textit{does} encode recordings from speakers that identify as men and women differently. 
 This aspect can serve as a proxy for measuring and mitigating gender disparities, e.g. by  reducing gender extractability from hidden representations \citep{krishnan2024encodinggendertransformerbasedasr}.
However, we caution the reader from attributing high extractability to gendered voice and discourage using probes on people's voice to predict gender (see Ethical considerations, \S\ref{sec:ethics}). 

\section{Discussion}
\label{sec:discussion}

Twists, humps, and pebbles---the way to equitable multilingual speech recognition models is no straight line. 
We have discovered that gender disparities vary across spoken setups and languages and that consistency is also weak between models. Perhaps of greater interest, the advantaged group is sometimes women, sometimes men, and never anyone who does not identify with either.

Such a high variability prevents us from foreseeing gender disparities and, in turn, the actual impact on system users.
We discuss here how to \textit{i}) reduce such variability and \textit{ii}) better use the metrics and test data at our disposal.

\paragraph{Free! the training data.}
While our acoustic and lexical inquiries on test data were inconsequential, experiments with probes on the model's internals showed that models learned and encoded properties to differentiate genders. Intuitively, a follow-up study would focus on what shaped those model's properties in the first place: training data. 
Models encode facts and information from training data, and some studies looked into how different sociodemographic groups are represented \citep{muller-etal-2023-gender,elazar2024whats}. 
Studying gender distributions in pretraining multilingual speech data would be extremely valuable, as we could establish how gendered gaps and underrepresentation correlate.
However, this analysis was impossible, as neither Whisper \citep{radford_robust_2022} nor SeamlessM4T \citep{communication2023seamless} released such information.
We call for more transparency.

\begin{figure}[!t]
    \centering
    \includegraphics[width=\linewidth]{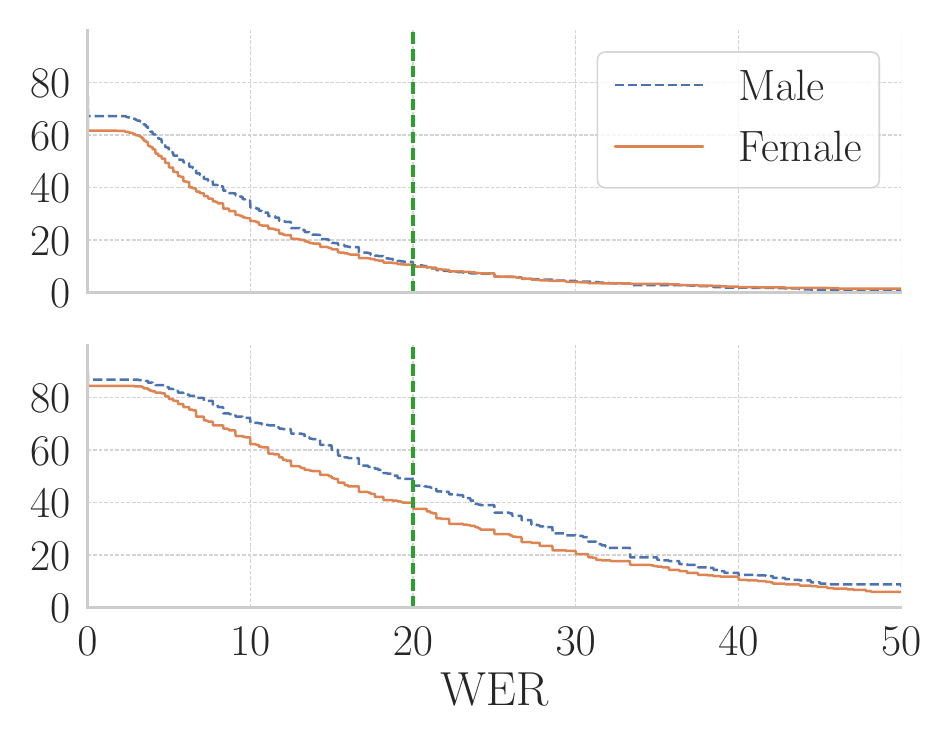}
    \caption{
    \textbf{Share of records (y axis, \%) having a WER greater
    than the value on the x axis.} \whisper, \fleurs nl (top) and ar (bottom).}
    \label{fig:sentence_eval}
\end{figure}

\paragraph{Go beyond group-wise metrics.}

A crucial choice in our design was measuring differences between two groups of speakers. We opted for $E(\cdot)$, measured as the \textit{relative} difference between the group means.
While the metric was primarily meant to make results across datasets, languages, and models comparable, it does not provide a complete picture of gender disparity's concrete impact. For instance, it hides \textit{absolute} gaps: a $E(\cdot)=10$ gap corresponds to a 0.2 WER gap if the model averages a WER of 2 on male speakers and to a 2 WER gap if the average is 20. These gaps may or may not be significant given many contextual factors (e.g., what the model is being used for, who is using it, and which ramifications such gaps can have). 
This thinking echoes that of recent critiques on measuring social bias in NLP, advocating for explicit statements about what system behavior must be considered good or bad and under which social values \citep{blodgett-etal-2020-language}, and how to build contextualized metrics \citep{lum2024bias}.


We would like a metric to indicate a more precise estimation of ASR model failures and their severity for users. 
A step in this direction would be going beyond group-wise metrics and studying sentence-level errors. Drawing on the concepts of wealth distribution and inequality \citep{o2003economics}, we can inspect error distributions between speakers and groups and address the question ``\textit{assuming X is an acceptable error rate, what fraction of records from each group will be served to a satisfactory level?}''  
Following \citet{koenecke2020}, we report in Figure \ref{fig:sentence_eval} the proportion of records having an error rate at least that large.
Due to space constraints, we focus on \whisper's transcriptions of \texttt{Fleurs}' Arabic and Dutch data.
This visualization lets us identify where performance between genders starts to diverge.
If, given our use case, we hypothesize that transcriptions become unusable for WER > 20, men and women would be equally affected in Dutch. However, it would not be the case in Arabic where 40\% of records from female speakers, but half of those from males, would not be acceptable. Parity is reached around WER=40.

This concrete analysis highlighted gender disparities that group-wise metrics could not capture otherwise. It also served as an example of a broader argumentation: Future inquiries can benefit from using metrics and observations more closely related to real-world scenarios. 




\paragraph{Improve sociodemographic representation.}
Empirical studies like ours rely 
on
the quality and quantity of evaluation data. To test 
speech recognition, we need representative splits 
from diverse speakers.
However, current datasets are gender-skewed, with men 
often 
overrepresented in terms of records and speakers.
While downsampling can help at evaluation time, it does not address the lack of speaker diversity.
Skewed distributions are even more evident in low-resource languages when building standard train/validation/test sets is required---e.g., in \cv, Yoruba counts only 14 and 20 unique males and females, respectively. Catalan counts only 8 within ``Other.''

Therefore, we call for much needed considerations when comparing gender groups in high-resource languages, and new collective efforts to collect more representative data for low-resource languages and gender identities 
beyond the binary.

\section{Conclusion}

We conducted the first extensive evaluation on gender-based performance gaps of Whisper \citep{radford_robust_2022} and SeamlessM4T \citep{communication2023seamless} for ASR. These models consistently exhibit gender bias across 19 languages from eight language families. Depending on the language, disparities can favor men or women but rarely favor speakers who identify as neither. We locate a potential source of these gaps using gender probes from interpretability approaches. Our results show that probes can be a proxy for gender gaps and that group fairness in multi-task multilingual models remains unsolved.

\section*{Ethical Considerations}
\label{sec:ethics}

The use of gender as a variable in this paper warrants ethical reflections.

As one of the most salient 
perceptual traits of one's identity \citep{kreiman2011foundations, azul2015varied, zimman2021gender}
gendered differences represent a linchpin of much (socio)phonetic research \citep{zimman2020sociophonetics}, which has unpacked several physical and sociocultural factors contributing to such gendered characteristics in the voice. Based on this evidence, our work does not intend to be normative nor assumes the existence of a single, unidimensional ``female'' or ``male'' voice. Instead, we question whether different gendered groups are equally recognized by current multilingual ASR models and incorporate sociophonetic knowledge in our analysis and discussions to isolate why that might not be the case. To do so, we do not make any inference about the gender of the speakers in the employed data. Instead, we rely on the declared gender of the speakers in the employed speech resources. In this regard, part of our analysis of the Mozilla Common Voice dataset uses a third category, ``Other'', which potentially aggregates diverse identities (e.g., transgender, non-binary, and other marginalized individuals)\footnote{These are among the offered gender options that can currently be reported when donating one's voice for Common Voice.} under one single umbrella term. While this third category also allows us to include genders non-conforming to the binary in our study, we recognize that this label and category might be an oversimplification and risk erasing the experiences and representativity of many gender identities.




Finally, gender probes represent a methodological approach to studying how models encode different audios from input signals. It is, hence, essential to remember the inherent limitations and ethical implications of relying on probes for gender classification. While these tools may offer a convenient means of categorization and can be suitable for exploring the models' behavior, they often overlook the nuanced and multifaceted nature of gender identity.

\section*{Limitations}

Our paper comes with a series of limitations. We divide them into two categories: data and methods.

\paragraph{Data.} Gender is the driving variable of our analysis. However, it is widely recognized that gender interacts with other sociocultural factors, e.g., dialect \citep{wolfram2004urban} or sexual orientation \citep{zimman2013hegemonic}, in voice production. By limiting to self-identified gender provided with the datasets, our analysis can only provide a partial view. We thus advocate for speech dataset releases that include a principled set of such factors. 

The scarcity of data limits the generalizability of our results in two more setups in our analysis. First, the category ``Other'' counts a deficient number of speakers and records in most of our setups (see \S\ref{sec:discussion}). Second, the number of speakers we extracted from \fleurs is 
 compared to \cv and \voxpopuli (see \ref{sec:eval}). These factors hamper cross-dataset comparison and overall generalizability.

We did not control for data quality on references. It might be that references are noisy (e.g., single words, empty) and can lead to over- or under-estimation of our measurements.

Finally, data contamination. Although we chose validation and test sets as our analysis targets, we cannot exclude that models were trained on part or all of them. We conducted a side analysis (Appendix~\ref{app:sampling_effect}) that suggested that this \textit{might} not be the case.

\paragraph{Methods.}

Since we miss reliable multilingual tools for acoustic analyses, we estimated the speaking rate using Whisper's pre-trained tokenizer. We acknowledge that pretrained tokenizers yield fewer tokens for high-resource languages. Hence, our measurement of low-resource might be overestimating the phenomenon.

If framed in the context of bias evaluation paradigms---which we are currently not doing---probing can be seen as an intrinsic bias paradigm. We correlate probing performance to a downstream task (and potential harm), i.e., quality gaps in ASR. However, studies in the field of NLP recognize that intrinsic and extrinsic (downstream) bias metrics do not necessarily correlate \citep{goldfarb-tarrant-etal-2021-intrinsic,kaneko-etal-2022-debiasing}. To our knowledge, ours is the first study to inspect such aspects in transformers for speech, and findings may not hold across modalities. We leave this research question to future work.      







\section*{Acknowledgments}

We thank the reviewers, the members of the SARDINE and MilaNLP research groups, and Marco Gaido for the insightful comments. Giuseppe Attanasio was supported by the Portuguese Recovery and Resilience Plan through project C645008882-00000055 (Center for Responsible AI) and by Fundação para a Ciência e Tecnologia through contract UIDB/50008/2020. He conducted part of the work as a member of the MilaNLP group at Bocconi University, Milan.
Dirk Hovy was supported by the European Research Council (ERC) under the European Union’s Horizon 2020 research and innovation program (grant agreement No.\ 949944, INTEGRATOR) and a MUR FARE 2020 initiative under grant agreement Prot.\ R20YSMBZ8S (INDOMITA). He is director of the Data and Marketing Insights Unit of the Bocconi Institute for Data Science and Analysis (BIDSA).
Beatrice Savoldi was supported by the PNRR project FAIR -  Future AI Research (PE00000013),  under the NRRP MUR program funded by the NextGenerationEU.

\bibliography{anthology,custom,probing}

\appendix

\section{Experimental Details}
\label{sec:appendix}

\begin{figure*}[t]
    \centering
    \begin{subfigure}{0.9\textwidth}
        \centering
        \includegraphics[width=\textwidth]{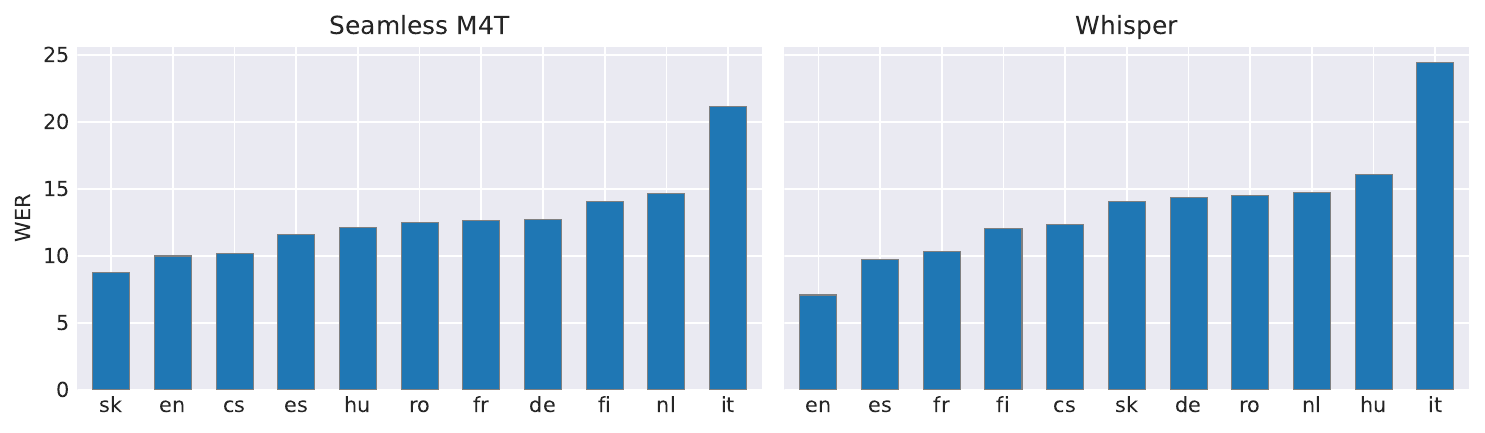}
        \caption{\voxpopuli}
        \label{fig:VPwer}
    \end{subfigure}
    
    \begin{subfigure}{0.9\textwidth}
        \centering
        \includegraphics[width=\linewidth]{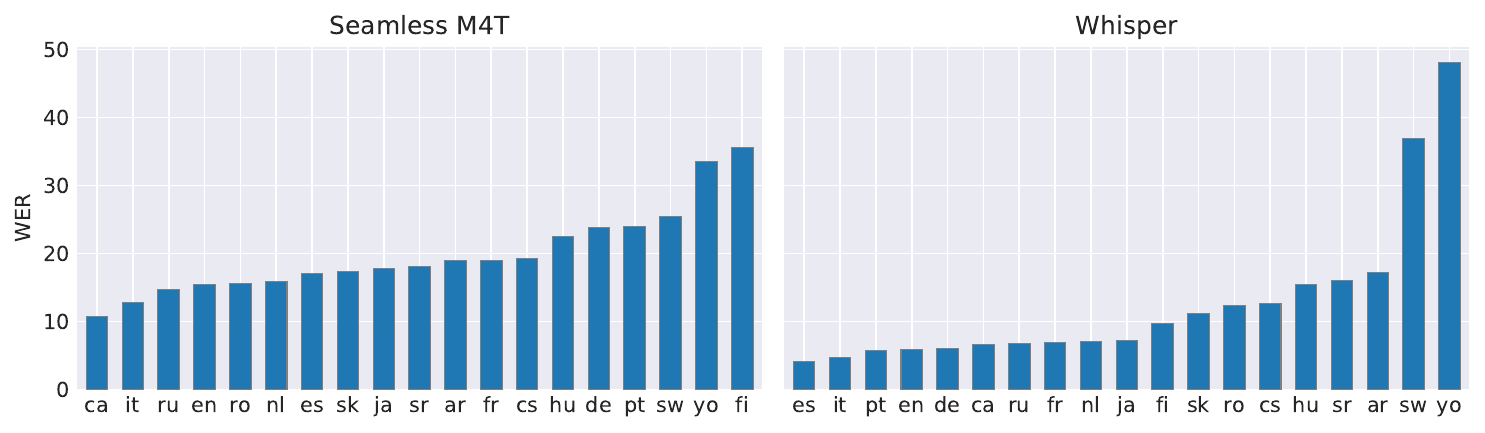}
        \caption{\fleurs}
        \label{fig:fleurswer}
    \end{subfigure}
    
    \begin{subfigure}{0.9\textwidth}
        \centering
        \includegraphics[width=\linewidth]{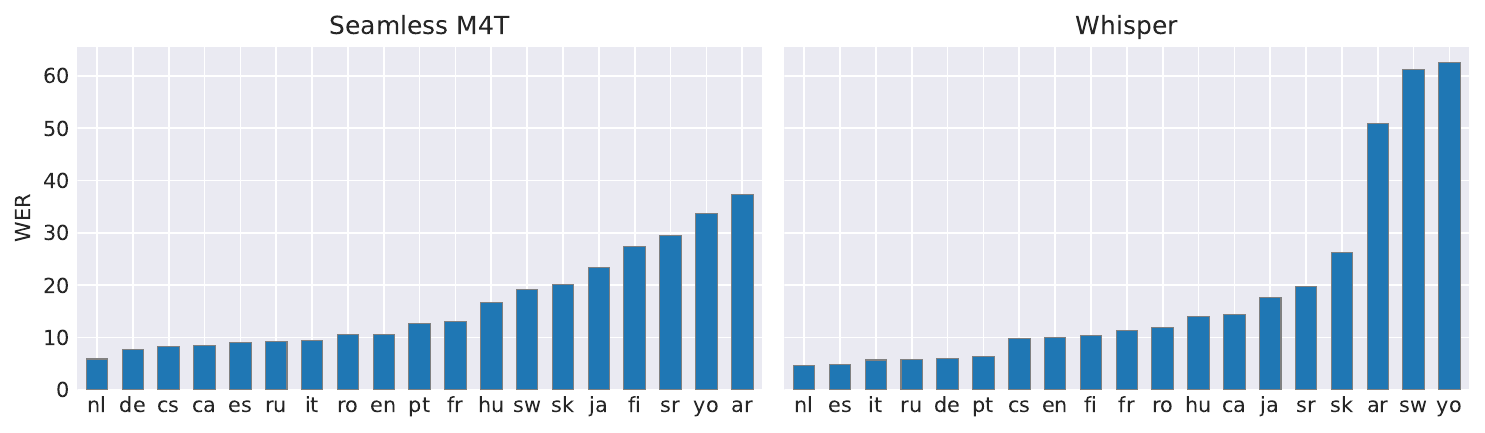}
        \caption{\cv}
        \label{fig:CVwer}
    \end{subfigure}
    \caption{\textbf{\seamless and \whisper transcription quality.} Error rate results are computed on test splits.}
    \label{fig:wer}
\end{figure*}

\subsection{Multilingual ASR Models}
\label{app:models}

For both Whisper and SeamlessM4T, we used code and model checkpoints in \texttt{transformers} \citep{wolf-etal-2020-transformers}. The Hub's model IDs are \href{https://huggingface.co/openai/whisper-large-v3}{openai/whisper-large-v3} and \href{https://huggingface.co/facebook/seamless-m4t-v2-large}{facebook/seamless-m4t-v2-large}, respectively. Both checkpoints correspond to the latest and best-performing versions available at the time of writing, February 2024.

We used each models's standard decoding configurations to transcribe each audio snippet.

In figure \ref{fig:wer}, we report WER results for \seamless and \whisper on the languages and datasets used in our experiments.


\begin{table*}[t]
\centering
\begin{tabular}{rlccccc} \toprule
& & \textbf{test} & \textbf{train} & \textbf{val} & \textbf{gender set} & \textbf{gender subset} \\ \midrule
\textbf{\texttt{Fleurs}} & Whisper & 12.68\footnotesize{${\pm.11.3}$} & 12.64\footnotesize{${\pm.11.4}$} & 12.49\footnotesize{${\pm.11}$} & 12.60\footnotesize{${\pm.11.2}$} & 12.62\footnotesize{${\pm.11.2}$} \\
& S M4T & 19.87\footnotesize{${\pm.6.4}$} & 19.72\footnotesize{${\pm.6.3}$} & 18.75\footnotesize{${\pm.6.2}$} & 19.65\footnotesize{${\pm.6.3}$} & 19.67\footnotesize{${\pm.6.3}$} \\
\midrule
\textbf{\texttt{CV}} & Whisper & 17.51\footnotesize{${\pm.17.5}$} & 16.65\footnotesize{${\pm.19.8}$} & 17.02\footnotesize{${\pm.18.7}$} & 16.88\footnotesize{${\pm.19.2}$} & 18.45\footnotesize{${\pm.19.2}$}   \\
& S M4T & 16.42\footnotesize{${\pm.9.5}$} & 12.68\footnotesize{${\pm9.9}$} & 14.7\footnotesize{${\pm.8.9}$} & 13.06\footnotesize{${\pm.9.6}$}&  15.58\footnotesize{${\pm.9.3}$}   \\
\midrule
\texttt{\textbf{VP}} & Whisper & 13.59\footnotesize{${\pm.4.45}$}  & 12.63\footnotesize{${\pm.3.79}$} & 12.99\footnotesize{${\pm.4.22}$} & 12.69\footnotesize{${\pm.3.8}$} & 13.32\footnotesize{${\pm.4.2}$} \\
& S M4T & 12.74\footnotesize{${\pm.3.27}$} & 11.27\footnotesize{${\pm.2.63}$} & 12.33\footnotesize{${\pm.2.68}$} & 11.39\footnotesize{${\pm.2.6}$} & 12.54\footnotesize{${\pm.2.9}$} \\
\bottomrule
\end{tabular}
\caption{\textbf{Error rate distribution} across datasets splits (test/train/validation), all pre-sampled female/male records, and on the sampled gender subset we used. Results averaged over 19 languages in \cv and \fleurs, and 11 for \voxpopuli.}
\label{tab:wer_splits}
\end{table*}


\begin{table*}[t]
\centering
\begin{tabular}{rlccccc} \toprule
& & test & train & val & presample\_all & overall \\
\midrule
\texttt{\textbf{Fleurs}} & Whisper & -1.18\footnotesize{${\pm.2.9}$} & -0.659\footnotesize{${\pm.1.8}$} & -0.255\footnotesize{${\pm.2.6}$} & -0.893\footnotesize{${\pm.1.7}$} & -0.997\footnotesize{${\pm.1.9}$} \\
& S M4T & 0.153\footnotesize{${\pm.1.9}$} & -0.880\footnotesize{${\pm.1}$} & -0.257\footnotesize{${\pm.2.7}$} & -0.735\footnotesize{${\pm.0.72}$} 
& -0.809\footnotesize{${\pm.0.72}$}  \\
\midrule
\texttt{\textbf{CV}} & Whisper & 1.17\footnotesize{${\pm.5.7}$} & -0.65\footnotesize{${\pm.2.7}$} & 0.88\footnotesize{${\pm.5.6}$} & -0.14\footnotesize{${\pm.3.5}$}  & 1.06\footnotesize{${\pm.5.3}$}  \\
& S M4T & 2.01\footnotesize{${\pm.5.4}$} & -1.22\footnotesize{${\pm.5.9}$} & 0.40\footnotesize{${\pm.2.7}$} & 0.59\footnotesize{${\pm.2.9}$} & 1.05\footnotesize{${\pm.3.3}$}   \\
\midrule
\texttt{\textbf{VP}} & Whisper & 0.31\footnotesize{${\pm.3.5}$} &  -0.80\footnotesize{${\pm.0.97}$} & -0.58\footnotesize{${\pm.1.9}$} & -0.73\footnotesize{${\pm.0.9}$} & -0.104\footnotesize{${\pm.2.2}$} \\
& S M4T & 0.16\footnotesize{${\pm.2.8}$} & -0.77\footnotesize{${\pm.0.54}$} & -0.71\footnotesize{${\pm.2.1}$} & -0.75\footnotesize{${\pm.0.6}$}
 & -0.268\footnotesize{${\pm.2.2}$} \\
\bottomrule
\end{tabular}
\caption{\textbf{Average F-M ER \textit{absolute} difference} across datasets splits (test/train/validation), all pre-sampled female/male records, and on the sampled gender subset used in our experiments. The results are averaged over 19 languages for \cv and 11 for \voxpopuli. For \fleurs, we average results over the only 9 languages comprising female/male speakers in all splits.}
\label{tab:wer_diff_splits}
\end{table*}

\subsection{Languages}

\begin{table}[!t]
\centering
\footnotesize
\begin{tabular}{@{}lll@{}}
\toprule
\textbf{Language} & \textbf{ISO} & \textbf{Family} \\ \midrule
Japanese & ja & Japonic \\ \midrule
Dutch & nl & Germanic \\
English & en &  \\
German & de &  \\ \midrule
Swahili & sw & NC / Bantu \\ \midrule
Yoruba & yo & NC / Volta-Niger \\ \midrule
Catalan & ca & Romance \\
French & fr &  \\
Italian & it &  \\
Portuguese & pt &  \\
Romanian & ro &  \\
Spanish & es &  \\ \midrule
Arab & ar & Semitic \\ \midrule
Czech & cs & Slavic \\
Russian & ru &  \\
Serbian & sr &  \\
Slovak & sk &  \\ \midrule
Finnish & fi & Uralic \\
Hungarian & hu &  \\ \bottomrule
\end{tabular}
\caption{\textbf{Languages}, their ISO code, and family studied in this paper.}
\label{tab:languages}
\end{table}

We studied 19 languages in \cv and \fleurs and 11 in \voxpopuli. Languages cover eight distinct language families. Due to data availability, we limited the comparison between speaker identifying with ``Male'' and ``Other'' in \cv to Catalan, German, English, Spanish, and French.
Table~\ref{tab:languages} reports an overview of languages and language families.

\subsection{Voice Activity Detection}
\label{app:vad}

While conducting our experiments, we noticed that other than records with an empty reference, datasets can contain also empty audio snippets. More precisely, these recordings do have a signal but it is mostly silence. We found this phenomenon primarily in Fleurs-es, with silence mostly coming from snippets attributed to women.  

Therefore, we used \texttt{pyannote.audio}'s pretrained neural models and code for voice activity detection.\footnote{https://github.com/pyannote/pyannote-audio} After counting the number of segments where voice was detected, we filter out all snippets were no segment was detected. We release the the counts and IDs of silence records in our repository.

\subsection{Speaker ID Attribution in \fleurs}
\label{app:fleurs_hdbscan}

\begin{figure*}[!t]
    \centering
    \includegraphics[width=.9\textwidth]{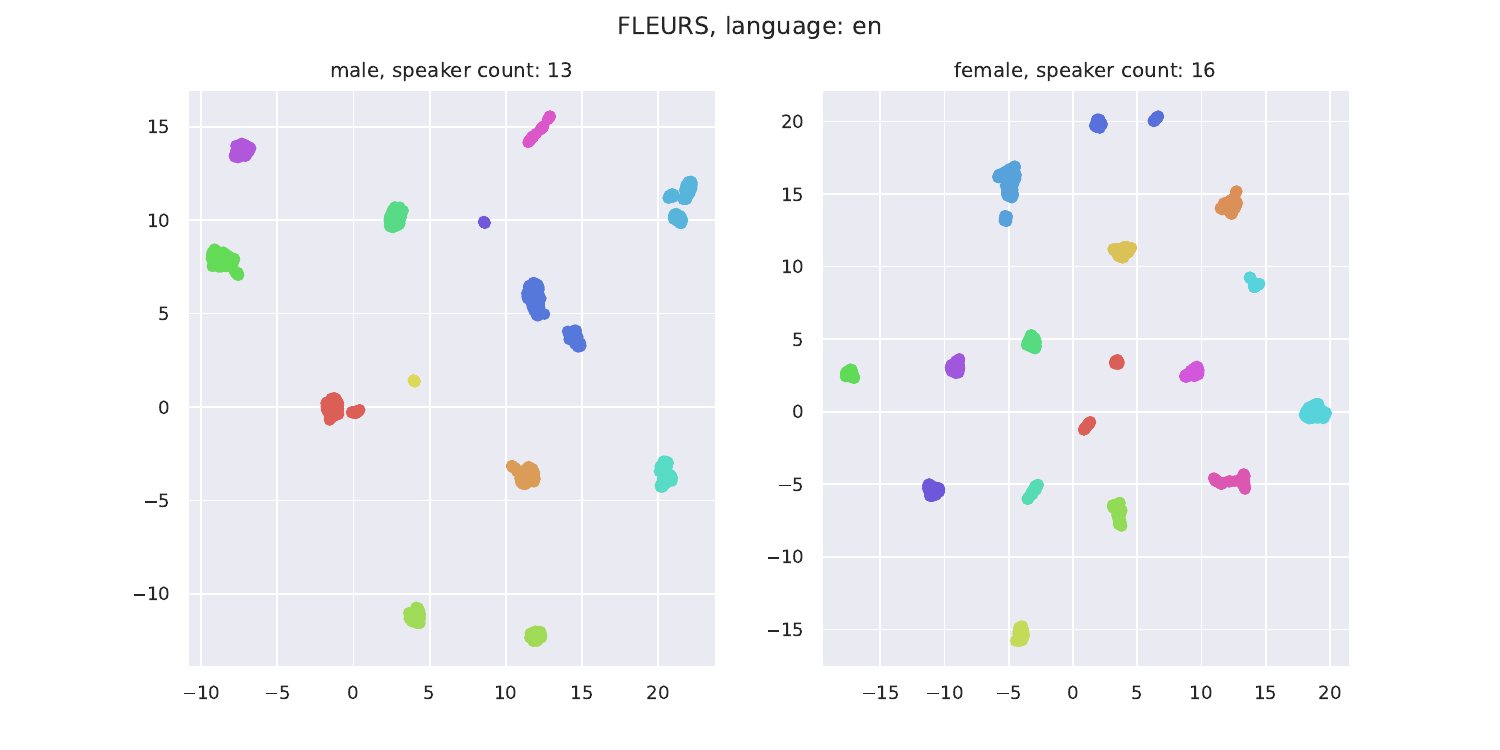}
    \caption{\textbf{UMAP projections of snippets in \fleurs-en.} We found 13 male speakers (left) and 16 female speakers (right). Color indicates cluster ID assigned by HDBSCAN.}
    \label{fig:umap_hdbscan}
\end{figure*}

Speaker IDs are crucial in our pipeline to avoid over- or under-estimating performance gaps due to overly present speakers.
Since \fleurs does not provide this piece of information, we attribute it automatically.

Specifically, we encode each snippet using a pretrained ECAPA-TDNN model for speaker verification. We use code from SpeechBrain \citep{speechbrain} and the model checkpoint with Hub ID \href{https://huggingface.co/speechbrain/spkrec-ecapa-voxceleb}{speechbrain/spkrec-ecapa-voxceleb}. Next, we cluster embeddings separately per gender using HDBSCAN \citep{campello2013density}, with code from \href{https://scikit-learn.org/stable/modules/generated/sklearn.cluster.HDBSCAN.html}{scikit-learn}. We use set \texttt{min\_cluster\_size} to 2 and use cosine similarity. Finally, we assign a numerical ID to each found cluster. Figure \ref{fig:umap_hdbscan} shows an example of clustered snippets projected with UMAP \citep{mcinnes2018umap}. We release per dataset and language IDs in our repository.

\subsection{Subset selection and data splits}
\label{app:sampling_effect}
To ensure the soundness of our experimental framework and of the gender subsample selection described in Section \ref{ssec:q1_exps}, we conducted preliminary analyses by comparing WER transcription quality scores (Table \ref{tab:wer_splits}) and gender F-M gap results (Figure \ref{tab:wer_diff_splits}) across: \textit{i)} different datasets splits (test/train/validation), \textit{ii)} all female/male records available from any split, and \textit{iii)} the sampled female/male subset used in our main experiments. 

Since no precise information concerning the data used to train \whisper and \seamless is publicly available, this evaluation is carried out with the goal of spotting potential data contamination issues. Namely, to ensure that we are not testing models on data samples comprised in their underlying training, thus posing the risk of obtaining unreliable results, which cannot be compared across datasets and models.

Table \ref{tab:wer_splits} shows no notable WER variations for \texttt{Fleurs}. Instead, both models perform better on the train splits of \texttt{CV} and \texttt{VP}, especially for the former dataset. The influence of the split accounted in \texttt{CV} and \texttt{VP} can thus be reflective of the F-M WER differences reported in Figure \ref{tab:wer_diff_splits}. Informed by such results, the gender subset used in our experiments is sampled from the test and validation split only for \texttt{CV} and \texttt{VP}, whereas for \texttt{Fleurs}, we leverage gender data from the whole corpus.

\subsection{Dataset Statistics}
\label{app:dataset_statistics}

\input{dataset_stats}

Table~\ref{tab:stats_cv}, \ref{tab:stats_fleurs}, and \ref{tab:stats_vp} reports all collected statistics on \cv, \fleurs, and \voxpopuli, respectively.

\subsection{Energy Statement}

Experiments were conducted using a private infrastructure on hardware of type A100 PCIe 80GB (TDP of 250W) with carbon efficiency of 0.233 kgCO2eq/kWh.\footnote{\url{https://app.electricitymaps.com}}
Consumption is mostly due to inference to generate transcripts.
We estimate a total emission of 3.6 kgCO2eq for experiments in the main body (transcripts of test and validation sets). Transcripts on training sets account for additional 104.24 kgCO2eq.
No emission was directly offset.
Total emissions are equivalent to driving an internal combustion engine car for 980 km.\footnote{Estimate based on average CO2 emissions of new passenger cars in EU in 2020. \url{https://www.acea.auto/figure/average-co2-emissions-of-new-cars-in-eu}}
Estimate was conducted using \texttt{codecarbon}\footnote{\url{https://github.com/mlco2/codecarbon}} but does not account for energy for cooling the infrastructure.

\section{Details on the exploratory analysis}

\begin{table}[t]
\small
\centering
\begin{tabular}{lllrr}
\toprule
\textbf{Dataset} & \textbf{Model} & \textbf{Feature} & \multicolumn{1}{c}{\textbf{\textit{rho}}} & \multicolumn{1}{c}{\textbf{\textit{p}}} \\
\midrule
\multirow{6}{*}{\voxpopuli} & \multirow{3}{*}{\seamless} & Sp. rate & -0.10 & 0.76 \\
                           &                           & Intensity     & -0.43 & 0.18 \\
                           &                           & Pitch         & \textbf{-0.83} & \textbf{0.00} \\
                           & \multirow{3}{*}{\whisper}  & Sp. rate & -0.33 & 0.32 \\
                           &                           & Intensity     & 0.23  & 0.50 \\
                           &                           & Pitch         & -0.30 & 0.36 \\ \hline
\multirow{6}{*}{\fleurs}    & \multirow{3}{*}{\seamless} & Sp. rate & \textbf{0.62}  & \textbf{0.00} \\
                           &                           & Intensity     & 0.11  & 0.66 \\
                           &                           & Pitch         & -0.06 & 0.82 \\
                           & \multirow{3}{*}{\whisper}  & Sp. rate & 0.20  & 0.42 \\
                           &                           & Intensity     & -0.26 & 0.29 \\
                           &                           & Pitch         & \textbf{-0.53} & \textbf{0.02} \\ \hline
\multirow{6}{*}{\cv}        & \multirow{3}{*}{\seamless} & Sp. rate & \textbf{-0.51} & \textbf{0.02} \\
                           &                           & Intensity     & -0.25 & 0.31 \\
                           &                           & Pitch         & 0.20  & 0.41 \\
                           & \multirow{3}{*}{\whisper}  & Sp. rate & -0.39 & 0.10 \\
                           &                           & Intensity     & -0.10 & 0.69 \\
                           &                           & Pitch         & 0.38  & 0.11 \\                 
\bottomrule
\end{tabular}
\caption{\textbf{Correlation between Acoustic Features and Error Rate Gaps}.
Pearson correlation between the difference of the mean of subgroups an  and $E(r_F,r_M)$. In bold the statistically relevant correlation ($p<0.05$).}
\label{tab:pearson}
\end{table}

\begin{table*}[!t]
\centering
\small
\begin{tabular}{llllllllllllllllllll}
\toprule 
                         & \multicolumn{1}{c}{\textbf{ar}} & \multicolumn{1}{c}{\textbf{ca}} & \multicolumn{1}{c}{\textbf{cs}} & \multicolumn{1}{c}{\textbf{de}} & \multicolumn{1}{c}{\textbf{en}} & \multicolumn{1}{c}{\textbf{es}} & \multicolumn{1}{c}{\textbf{fi}} & \multicolumn{1}{c}{\textbf{fr}} & \multicolumn{1}{c}{\textbf{hu}} & \multicolumn{1}{c}{\textbf{it}} & \multicolumn{1}{c}{\textbf{ja}} & \multicolumn{1}{c}{\textbf{nl}} & \multicolumn{1}{c}{\textbf{pt}} & \multicolumn{1}{c}{\textbf{ro}} & \multicolumn{1}{c}{\textbf{ru}} & \multicolumn{1}{c}{\textbf{sk}} & \multicolumn{1}{c}{\textbf{sr}} & \multicolumn{1}{c}{\textbf{sw}} & \multicolumn{1}{c}{\textbf{yo}} \\ \midrule
                         & \multicolumn{19}{c}{\textbf{\fleurs}} \\
\textbf{Intensity} & \checkmark                        & \checkmark                        & \checkmark                        & \checkmark                        & \checkmark                        & \checkmark                        & \checkmark                        & \checkmark                        & \checkmark                        & \checkmark                        & \checkmark                        & \checkmark                        & \checkmark                        & \checkmark                        & \checkmark                        & \checkmark                        & \checkmark                        & \checkmark                        & \checkmark                        \\
\textbf{Pitch}     & \checkmark                        & \checkmark                        & \checkmark                        & \checkmark                        & \checkmark                        & \checkmark                        & \checkmark                        & \checkmark                        & \checkmark                        & \checkmark                        & \checkmark                        & \checkmark                        & \checkmark                        & \checkmark                        & \checkmark                        & \checkmark                        & \checkmark                        & \checkmark                        & \checkmark                        \\
\textbf{Speaking rate}  & \checkmark                        & \checkmark                        & \checkmark                        & \checkmark                        & $\times$       & \checkmark                        & \checkmark                        & \checkmark                        & \checkmark                        & \checkmark                        & \checkmark                        & $\times$       & \checkmark                        & \checkmark                        & \checkmark                        & \checkmark                        & \checkmark                        & \checkmark                        & \checkmark                        \\  \midrule
                         & \multicolumn{19}{c}{\textbf{\voxpopuli}}                                                                                                                                                                                                                                                                                                                                                                                                                                                                                                                                                                                                                             \\
\textbf{Intensity} &                                 &                                 & \checkmark                        & \checkmark                        & $\times$       & \checkmark                        & \checkmark                        & \checkmark                        & \checkmark                        & $\times$       &                                 & \checkmark                        &                                 & \checkmark                        &                                 & \checkmark                        &                                 &                                 &                                 \\
\textbf{Pitch}     &                                 &                                 & \checkmark                        & \checkmark                        & \checkmark                        & \checkmark                        & \checkmark                        & \checkmark                        & \checkmark                        & \checkmark                        &                                 & \checkmark                        &                                 & \checkmark                        &                                 & \checkmark                        &                                 &                                 &                                 \\
\textbf{Speaking rate}  &                                 &                                 & \checkmark                        & \checkmark                        & \checkmark                        & \checkmark                        & $\times$       & $\times$       & \checkmark                        & \checkmark                        &                                 & $\times$       &                                 & \checkmark                        &                                 & \checkmark                        &                                 &                                 &                                 \\ \midrule
                         & \multicolumn{19}{c}{\textbf{\cv}}                                                                                                                                                                                                                                                                                                                                                                                                                                                                                                                                                                                                                                    \\
\textbf{Intensity} & $\times$       & \checkmark                        & \checkmark                        & \checkmark                        & \checkmark                        & \checkmark                        & \checkmark                        & \checkmark                        & $\times$       & \checkmark                        & \checkmark                        & \checkmark                        & \checkmark                        & \checkmark                        & \checkmark                        & $\times$       & \checkmark                        & \checkmark                        & \checkmark                        \\
\textbf{Pitch}     & \checkmark                        & \checkmark                        & \checkmark                        & \checkmark                        & \checkmark                        & \checkmark                        & \checkmark                        & \checkmark                        & \checkmark                        & \checkmark                        & \checkmark                        & \checkmark                        & \checkmark                        & \checkmark                        & \checkmark                        & \checkmark                        & \checkmark                        & \checkmark                        & \checkmark                        \\
\textbf{Speaking rate}  & \checkmark                        & $\times$       & \checkmark                        & \checkmark                        & $\times$       & $\times$       & \checkmark                        & $\times$       & \checkmark                        & \checkmark                        & \checkmark                        & \checkmark                        & $\times$       & \checkmark                        & \checkmark                        & \checkmark                        & \checkmark                        & \checkmark                        & \checkmark                        \\ \bottomrule
\end{tabular}
\caption{\textbf{Statistical differences of acoustic features between genders}, separately by language and dataset. 
\checkmark indicates statistical difference, $\times$ indicates no difference (independent-sample Student's T-test, $p<0.05$).}
\label{tab:ttest}
\end{table*}

\begin{figure*}[!t]
\centering
  \begin{subfigure}{0.7\textwidth}
    \centering
    \includegraphics[width=\linewidth]{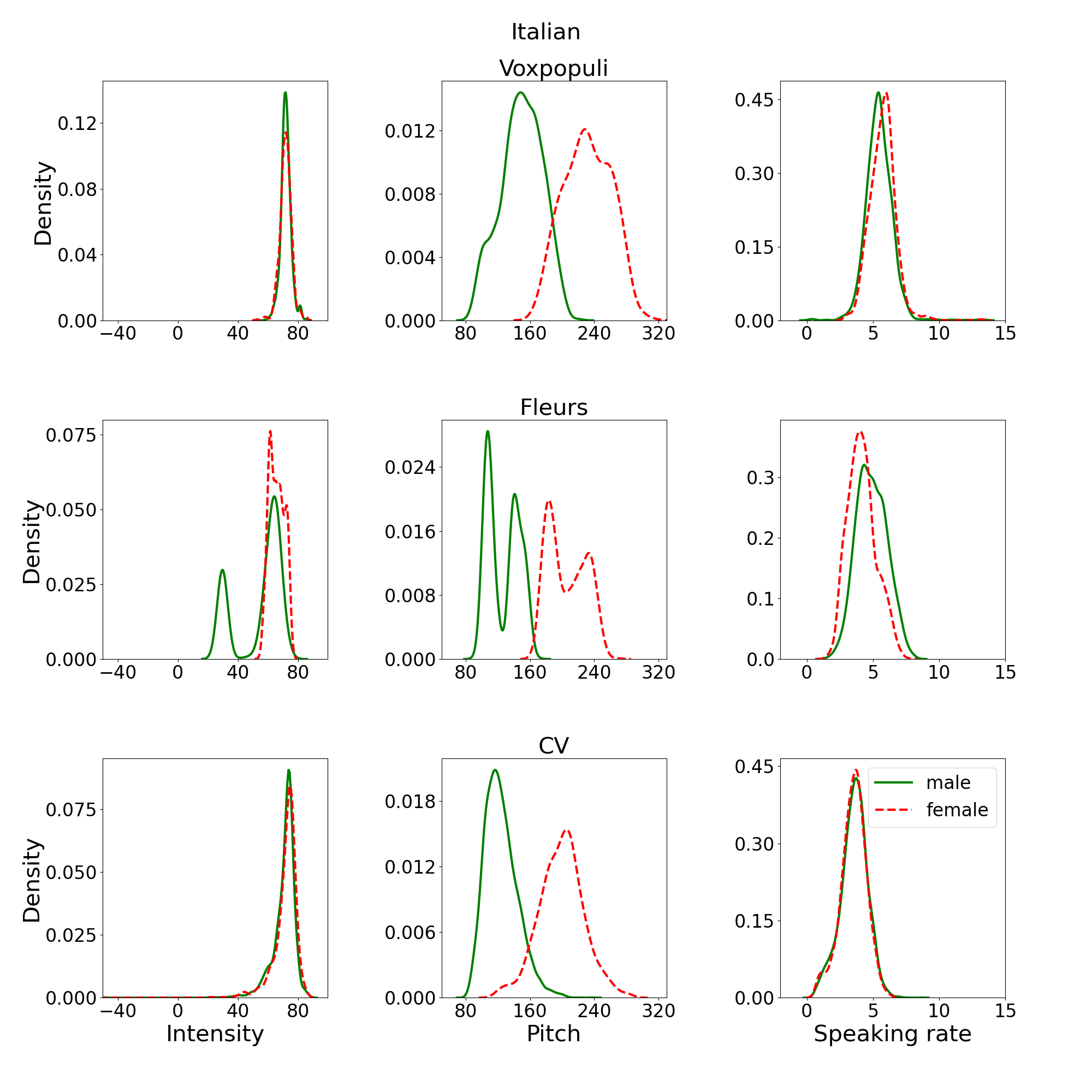}
  \end{subfigure}
  \hspace{0.04\textwidth}
  \begin{subfigure}{0.7\textwidth}
    \centering
    \includegraphics[width=\linewidth]{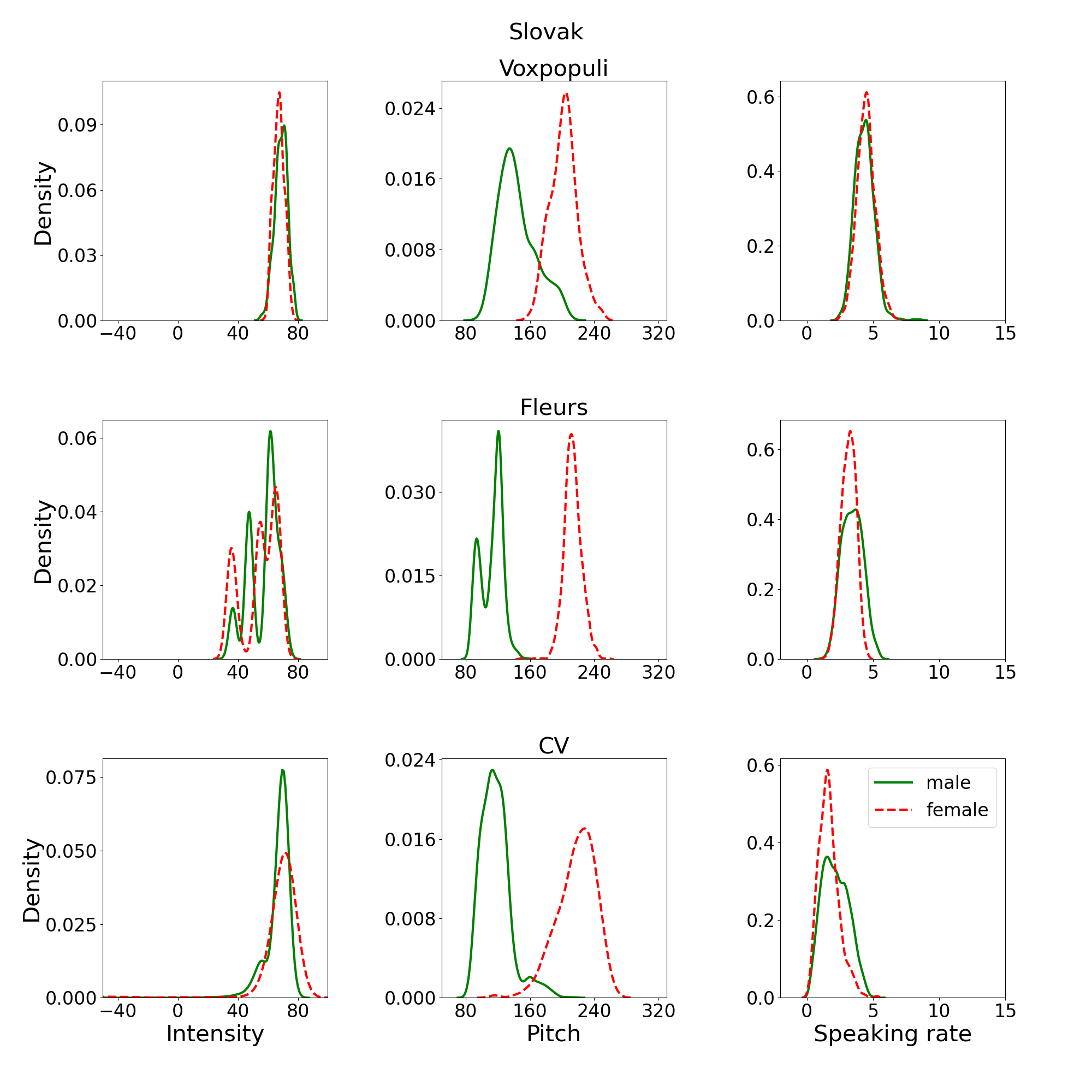}
  \end{subfigure}
  \caption{\textbf{Acoustic features.} Distributions of mean intensity, mean pitch, and speaking rate for Italian and Slovak on our three datasets. ``Male'' (green, solid line), ``Female'' (red, dashed line).}
  \label{fig:acoustic-features}
\end{figure*}

\subsection{Acoustic Analysis}
\label{app:ssec:acoustic_analysis}

We provide further details about the acoustic analysis carried out in \S \ref{sec:acoustic}, comparing male and female speakers.
Specifically, Table \ref{tab:ttest} reports the statistics of the T-tests conducted between acoustic features comparing genders across different languages and datasets, and Table \ref{tab:pearson} presents statistics on the Pearson correlation between differences in acoustic features and error rate gaps, separately for each (dataset, model) configuration.

We discuss the results of two representative languages only: Italian and Slovak. 
Italian exemplifies languages with comparable performance across genders, whereas Slovak is biased towards men and belongs to the set of languages with larger gender gap trends (see \S\ref{ssec:q1_results} for details).
For both languages, acoustic features are, in most cases, statistically different. As shown in Figure \ref{fig:acoustic-features}, the most significant differences are found in mean pitch values, confirming the well-known differences between males and females. The distributions align with ranges suggested by the literature \citep{simpson2009phonetic} for both languages and all datasets.
In contrast, differences in speaking rate are much less evident but still statistically significant according to the T-test (see Table \ref{tab:ttest}). Intensity shows the most variability across all dimensions, even within gender groups, particularly noticeable in the \texttt{Fleurs} dataset.\footnote{This variability could be attributed to the underlying recording conditions of \texttt{Fleurs}, details of which were not provided in the original paper.} In Italian (\texttt{VoxPopuli}), there is no difference in intensity between the two gender groups ($p=0.356$), and similarly in Slovak (\texttt{CV}) ($p=0.376$).
However, for both languages, fitting an OLS regression at the sentence level using acoustic features to predict sentence-level error rates ($r_F$, $r_M$) showed no significant contribution, similar to other languages (R$^2$ with 
max: $0.20$ and average$_\sigma$: $0.03_{\pm0.04}$)
 
%
%

\subsection{Lexical Analysis}
\label{app:lexical_analysis}

To capture lexical phenomena, we extracted from the reference transcript of each record two sets of features. \textbf{Part-of-speech} tags include NOUN, PROPN, VERB, ADJ, ADV, PRON, AUX, CCONJ, DET, PART. \textbf{Named entities} tags include F\_LOC, F\_ORG, F\_PER, and MISC. For each language, we used the corresponding SpaCy (\url{https://spacy.io/models}) ``medium'' text analyzer (e.g., for English \texttt{en\_core\_web\_sm} and \texttt{de\_core\_news\_sm} for German) when available. For \texttt{hu} an \texttt{ar} we used models available in Stanza \citep{qi-etal-2020-stanza}. For \texttt{yo} and \texttt{sw} we limited the analysis to POS tags due to NE parses unavailability. We used \url{mbeukman/xlm-roberta-base-finetuned-ner-yorub} and \url{mbeukman/xlm-roberta-base-finetuned-swahili-finetuned-ner-swahili} for \texttt{yo} and \texttt{sw}, respectively \citep{beukman-fokam-2023-analysing}. 
We left \texttt{sr}, \texttt{sk}, and \texttt{cs} out of this analysis since no taggers were available. 
We normalized tags from different modes accordingly (e.g., we consider Stanza's PERSON and SpaCy F\_PER the same feature).
Moreover, we computed lexical complexity as specified in \citet{Imani2017LexicalFO}, i.e., the ratio between the sum of [NOUN, PROPN, VERB, ADJ, ADV] and the sum of [PRON, AUX, CCONJ, DET, PART].

\subsection{Probing Analysis}
\label{app:probing_details}

Training and test set for probing experiments are sampled from the concatenation of each language's validation and test sets for consistency with error rate experiments. We balanced the samples on gender, stratifying on the speaker distribution.

We used standard logistic regression probes as per \texttt{scikit-learn}'s implementation and standard parameters but a larger allowance of steps for convergence (n=1000). For MDL probes, we used logistic regression as the backbone classifier for comparability and operationalized the probe via the \textit{online code} configuration \cite{voita-titov-2020-information}. Following previous work \citep{orgad-etal-2022-gender}, we set the dataset slices to (percentage): \texttt{[0.2, 0.4, 0.8, 1.6, 3.2, 6.25, 12.5, 25, 50, 100]}.

We probe a total of 1000 distinct positions that correspond roughly to the first tens seconds of each recording (Whisper encodes audio with 25 milliseconds-long rolling frame and stride of 10 milliseconds) \citep{radford_robust_2022}. 

\section{Release Statement}

The following is a list of artifacts we produced in this work that we release to facilitate future research.
\begin{itemize}
    \item Transcriptions obtained with Whisper and SeamlessM4T of the full datasets Mozilla Common Voice, Google Fleurs, and Meta VoxPopuli.
    \item Segment-level annotations of segments with voice activity extracted from the voice activity detection pipeline (\S\ref{app:vad}).
    \item Segment-level acoustic features from our acoustic analysis 
    (\S\ref{sec:acoustic}).
    \item Segment-level embeddings and cluster IDs extracted on Fleurs with the speaker identification pipeline (\S\ref{app:fleurs_hdbscan}).
    \item Extensive statistics on speaker, gender, and record distributions on the three datasets---which we could not find anywhere else online.
    \item Dataset samples on which we computed the quality and gap metrics for reproducibility purposes (\S\ref{ssec:q1_exps}). 
\end{itemize}

\begin{figure*}
    \centering

    \begin{subfigure}{\textwidth}
        \centering
        \includegraphics[width=.95\linewidth]{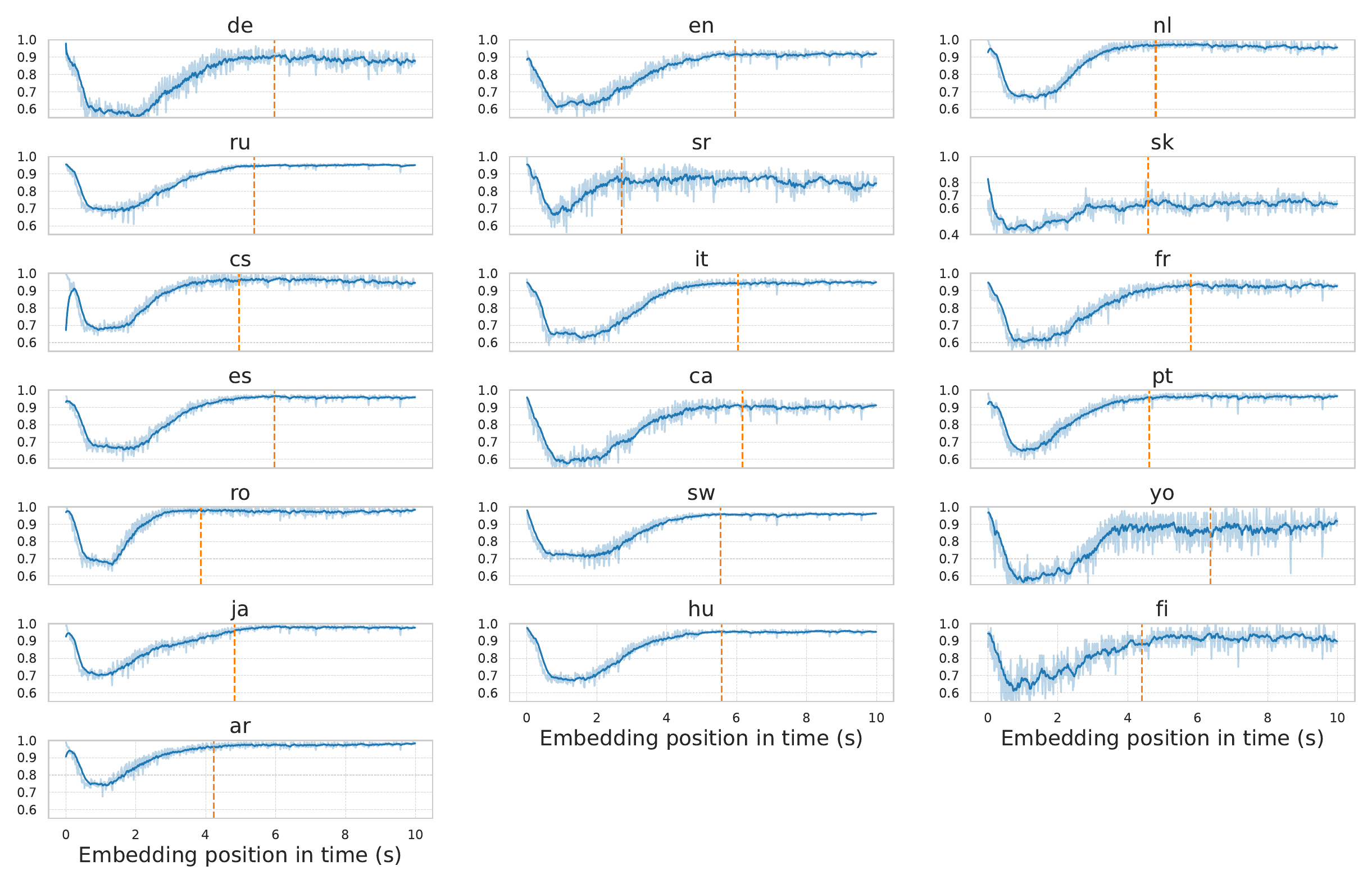}
        \caption{Probes trained on the \textit{original} labels.}
        \label{sfig:logreg_fm_normal}
    \end{subfigure}

    \begin{subfigure}{\textwidth}
        \centering
        \includegraphics[width=.95\linewidth]{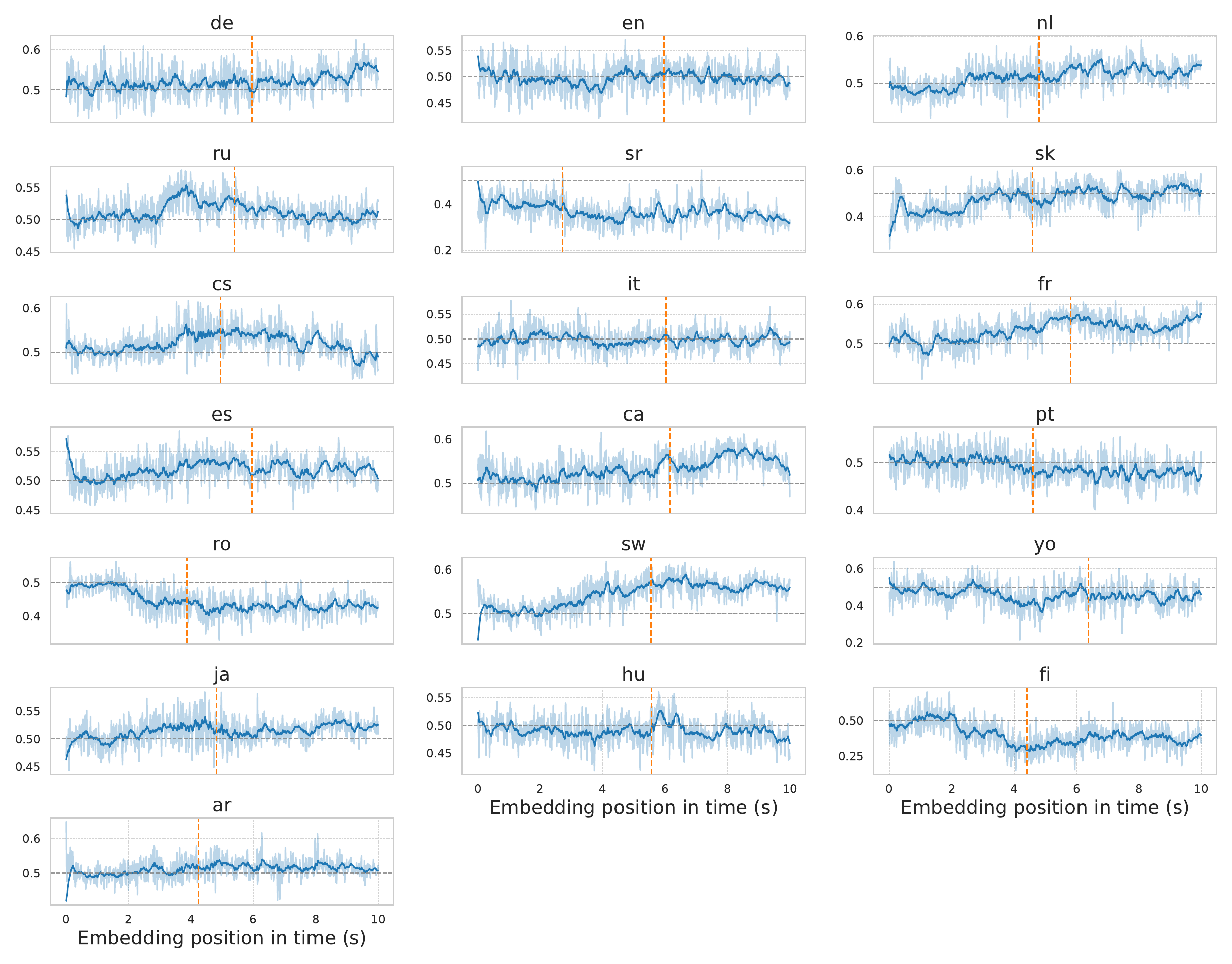}
        \caption{Probes trained on the \textit{shuffled} labels.}
        \label{sfig:logreg_fm_random}
    \end{subfigure}

    \caption{F-M gender probing F1 Macro performance for every contextual embedding in Whisper-large-v3 within the first 10 seconds (x axis). Logistic regression probe with L2 regularization trained on standard (\subref{sfig:logreg_fm_normal}) and shuffled (\subref{sfig:logreg_fm_random}) training labels. Orange lines indicate the average length of test segments. All \cv languages.}
\end{figure*}

\end{document}

%% file: dataset_stats.tex

\newcommand{\foot}[1]{\footnotesize{$\pm#1$}}

\begin{table*}[!t]
\footnotesize
\centering
\begin{tabular}{lrrrrrrrrrr}\toprule
\textbf{Lang} &\textbf{\# records} &\textbf{Seconds} &\textbf{\# Tokens} &\textbf{\# M} &\textbf{Gini (M)} &\textbf{\# F} &\textbf{Gini (F)} & \textbf{\# O} &\textbf{Gini (O)} \\\midrule
ar &14013 &4.37\foot{1.51} &19.92\foot{17.55} &319 &0.66 &102 &0.68 \\
ca &1645 &6.12\foot{1.73} &17.05\foot{6.22} &265 &0.22 &241 &0.22 & 8 & 0.26 \\
cs &13746 &4.60\foot{1.38} &20.77\foot{8.71} &293 &0.64 &39 &0.61 \\
de &3641 &6.10\foot{1.75} &16.74\foot{6.08} &807 &0.22 &129 &0.24 & 17 & 0.24\\
en &4938 &5.96\foot{2.28} &11.08\foot{3.99} &2262 &0.25 &476 &0.25 & 33 & 0.21 \\
es &6404 &6.09\foot{1.60} &15.61\foot{5.01} &1476 &0.29 &517 &0.30 & 42 & 0.34 \\
fi &1695 &4.74\foot{1.66} &18.22\foot{8.93} &52 &0.57 &17 &0.57 \\
fr &4214 &5.81\foot{1.67} &16.40\foot{5.56} &876 &0.19 &213 &0.19 & 19 & 0.23 \\
hu &12245 &5.45\foot{1.52} &22.84\foot{9.23} &173 &0.57 &277 &0.53 \\
it &7088 &6.14\foot{1.69} &17.52\foot{6.44} &888 &0.29 &203 &0.28 \\
ja &7981 &5.02\foot{2.29} &22.64\foot{15.07} &1033 &0.44 &632 &0.41 \\
nl &12808 &4.83\foot{1.38} &16.29\foot{5.57} &487 &0.62 &130 &0.63 \\
pt &7319 &4.61\foot{1.63} &10.66\foot{5.44} &620 &0.41 &82 &0.37 \\
ro &6489 &4.05\foot{0.90} &15.70\foot{4.13} &167 &0.56 &39 &0.57 \\
ru &10466 &5.51\foot{1.85} &18.87\foot{9.18} &702 &0.45 &248 &0.43 \\
sk &3008 &4.34\foot{1.74} &13.66\foot{9.64} &58 &0.64 &13 &0.65 \\
sr &1895 &2.86\foot{0.89} &8.63\foot{5.97} &39 &0.57 &11 &0.64 \\
sw &13730 &5.61\foot{1.80} &21.04\foot{8.13} &248 &0.62 &251 &0.58 \\
yo &1226 &5.77\foot{1.40} &41.97\foot{9.89} &14 &0.48 &20 &0.64 \\
\bottomrule
\end{tabular}

\caption{\textbf{Statistic for \cv}. We used validation and test sets. Total number of records, average length in seconds, number of tokens as tokenized by Whisper's pretrained tokenizer, number of unique speakers and, gini index of the snippets-per-speaker dispersion for male (M), female (F), and ``other'' (O) subgroups.}
\label{tab:stats_cv}

\end{table*}

\begin{table*}[!t]
\footnotesize
\centering
\begin{tabular}{lrrrrrrrr}\toprule
\textbf{Lang} &\textbf{\# records} &\textbf{Seconds} &\textbf{\# Tokens} &\textbf{\# M} &\textbf{Gini (M)} &\textbf{\# F} &\textbf{Gini (F)} \\\midrule
cs &2208 &9.93\foot{5.38} &60.04\foot{31.51} &45 &0.58 &16 &0.47 \\
de &4064 &8.74\foot{5.87} &32.34\foot{20.82} &147 &0.55 &64 &0.57 \\
en &3485 &9.92\foot{6.17} &24.48\foot{15.05} &274 &0.52 &104 &0.52 \\
es &3130 &11.56\foot{6.78} &42.89\foot{25.15} &95 &0.57 &48 &0.48 \\
fi &1141 &9.99\foot{5.44} &46.65\foot{26.61} &25 &0.63 &16 &0.45 \\
fr &3341 &10.23\foot{6.15} &38.19\foot{23.92} &151 &0.52 &84 &0.54 \\
hu &2145 &10.47\foot{6.05} &62.48\foot{35.59} &45 &0.48 &21 &0.50 \\
it &2419 &13.00\foot{7.19} &52.78\foot{28.99} &87 &0.47 &37 &0.53 \\
nl &2351 &8.16\foot{5.23} &36.63\foot{22.88} &69 &0.54 &42 &0.54 \\
ro &2701 &11.29\foot{5.75} &57.31\foot{31.54} &51 &0.52 &19 &0.57 \\
sk &1266 &10.33\foot{5.55} &62.52\foot{35.16} &26 &0.57 &10 &0.47 \\ 
\bottomrule
\end{tabular}

\caption{\textbf{Statistic for \voxpopuli}. We used validation and test sets. Total number of records, average length in seconds, number of tokens as tokenized by Whisper's pretrained tokenizer, number of unique speakers and, gini index of the snippets-per-speaker dispersion for male (M) and female (F) subgroups.
}\label{tab:stats_vp}

\end{table*}

\begin{table*}[!t]
\footnotesize
\centering
\begin{tabular}{lrrrrrrrr}\toprule
\textbf{Lang} &\textbf{\# records} &\textbf{Seconds} &\textbf{\# Tokens} &\textbf{\# M} &\textbf{Gini (M)} &\textbf{\# F} &\textbf{Gini (F)} \\\midrule
ar &836 &10.83\foot{4.08} &57.52\foot{21.70} &9 &0.31 &38 &0.51 \\
ca &2914 &11.87\foot{4.00} &39.93\foot{14.87} &5 &0.15 &5 &0.05 \\
cs &3772 &12.05\foot{4.45} &53.38\foot{19.93} &7 &0.36 &5 &0.21 \\
de &3326 &12.98\foot{6.32} &37.19\foot{13.30} &4 &0.07 &4 &0.47 \\
en &2832 &9.75\foot{3.58} &24.55\foot{8.54} &13 &0.33 &15 &0.28 \\
es &3082 &12.16\foot{3.80} &37.88\foot{13.67} &6 &0.38 &7 &0.35 \\
fi &3482 &12.74\foot{4.43} &47.38\foot{18.27} &12 &0.34 &18 &0.47 \\
fr &3194 &10.25\foot{3.58} &38.80\foot{15.06} &4 &0.06 &6 &0.15 \\
hu &4374 &12.09\foot{4.12} &54.90\foot{21.68} &5 &0.03 &5 &0.05 \\
it &3528 &14.53\foot{5.34} &40.59\foot{14.30} &6 &0.33 &6 &0.31 \\
ja &2348 &12.99\foot{3.77} &49.50\foot{18.25} &6 &0.50 &16 &0.66 \\
nl &2594 &9.55\foot{3.73} &40.59\foot{14.63} &8 &0.53 &6 &0.23 \\
pt &3944 &12.50\foot{4.24} &35.32\foot{12.88} &5 &0.08 &4 &0.19 \\
ro &4022 &10.22\foot{3.63} &48.98\foot{18.03} &5 &0.08 &7 &0.35 \\
ru &3496 &11.39\foot{3.97} &42.57\foot{16.14} &13 &0.36 &11 &0.26 \\
sk &2600 &11.65\foot{3.81} &53.15\foot{19.93} &8 &0.41 &10 &0.57 \\
sr &2820 &10.76\foot{3.40} &59.46\foot{21.59} &4 &0.07 &6 &0.08 \\
sw &3588 &14.08\foot{4.52} &49.60\foot{17.25} &18 &0.42 &17 &0.39 \\
yo &2850 &16.32\foot{5.66} &73.91\foot{34.05} &7 &0.36 &6 &0.50 \\
\bottomrule
\end{tabular}

\caption{\textbf{Statistic for \fleurs.} We used validation and test sets. Total number of records, average length in seconds, number of tokens as tokenized by Whisper's pretrained tokenizer, number of unique speakers and, gini index of the snippets-per-speaker dispersion for male (M) and female (F) subgroups. Speakers are extracted automatically.
}\label{tab:stats_fleurs}

\end{table*}